%% file: paper.tex
\documentclass[]{bytedance_seed}

\usepackage[toc,page,header]{appendix}

\input{resources/packages}

\input{resources/math_macro}
\usepackage{mathrsfs}
\usepackage{adjustbox}
\usepackage{multirow}
\usepackage{multirow}
\usepackage{multicol}
\usepackage{tcolorbox}
\usepackage{changepage}
\usepackage{graphicx}
\usepackage{amssymb}
\usepackage{array}
\usepackage{bm}

\usepackage{minitoc}

\newcommand{\dif}{\mathrm d}

\title{Seedream 3.0 Technical Report}

\author[]{ByteDance Seed}

\abstract{
We present Seedream 3.0, a high-performance Chinese-English bilingual image generation foundation model. We develop several technical improvements to address existing challenges in Seedream 2.0, including alignment with complicated prompts, fine-grained typography generation, suboptimal visual aesthetics and fidelity, and limited image resolutions. Specifically, the advancements of Seedream 3.0 stem from improvements across the entire pipeline, from data construction to model deployment. At the data stratum, we double the dataset using a defect-aware training paradigm and a dual-axis collaborative data-sampling framework. Furthermore, we adopt several effective techniques such as mixed-resolution training, cross-modality RoPE, representation alignment loss, and resolution-aware timestep sampling in the pre-training phase. During the post-training stage, we utilize diversified aesthetic captions in SFT, and a VLM-based reward model with scaling, thereby achieving outputs that well align with human preferences. Furthermore, Seedream 3.0 pioneers a novel acceleration paradigm. By employing consistent noise expectation and importance-aware timestep sampling, we achieve a 4 to 8 times speedup while maintaining image quality. Seedream 3.0 demonstrates significant improvements over Seedream 2.0: it enhances overall capabilities, in particular for text-rendering in complicated Chinese characters which is important to professional typography generation. In addition, it provides native high-resolution output (up to 2K), allowing it to generate images with high visual quality. Seedream 3.0 is now accessible on \href{https://console.volcengine.com/ark/region:ark+cn-beijing/experience/vision?type=GenImage}{Volcano Engine}\textsuperscript{$\alpha$}.
}

\checkdata[Official Page]{\url{https://team.doubao.com/tech/seedream3_0} }
\checkdata[\textsuperscript{$\alpha$}Model ID]{Doubao-Seedream-3.0-t2i}

\begin{document}
\begin{CJK*}{UTF8}{gbsn}

\maketitle

\definecolor{chinese_red}{HTML}{8B4513}
\definecolor{english_blue}{HTML}{4169E1}

\begin{figure}[ph]
\begin{center}
\vspace{-30pt}
\includegraphics[height=5.6cm]{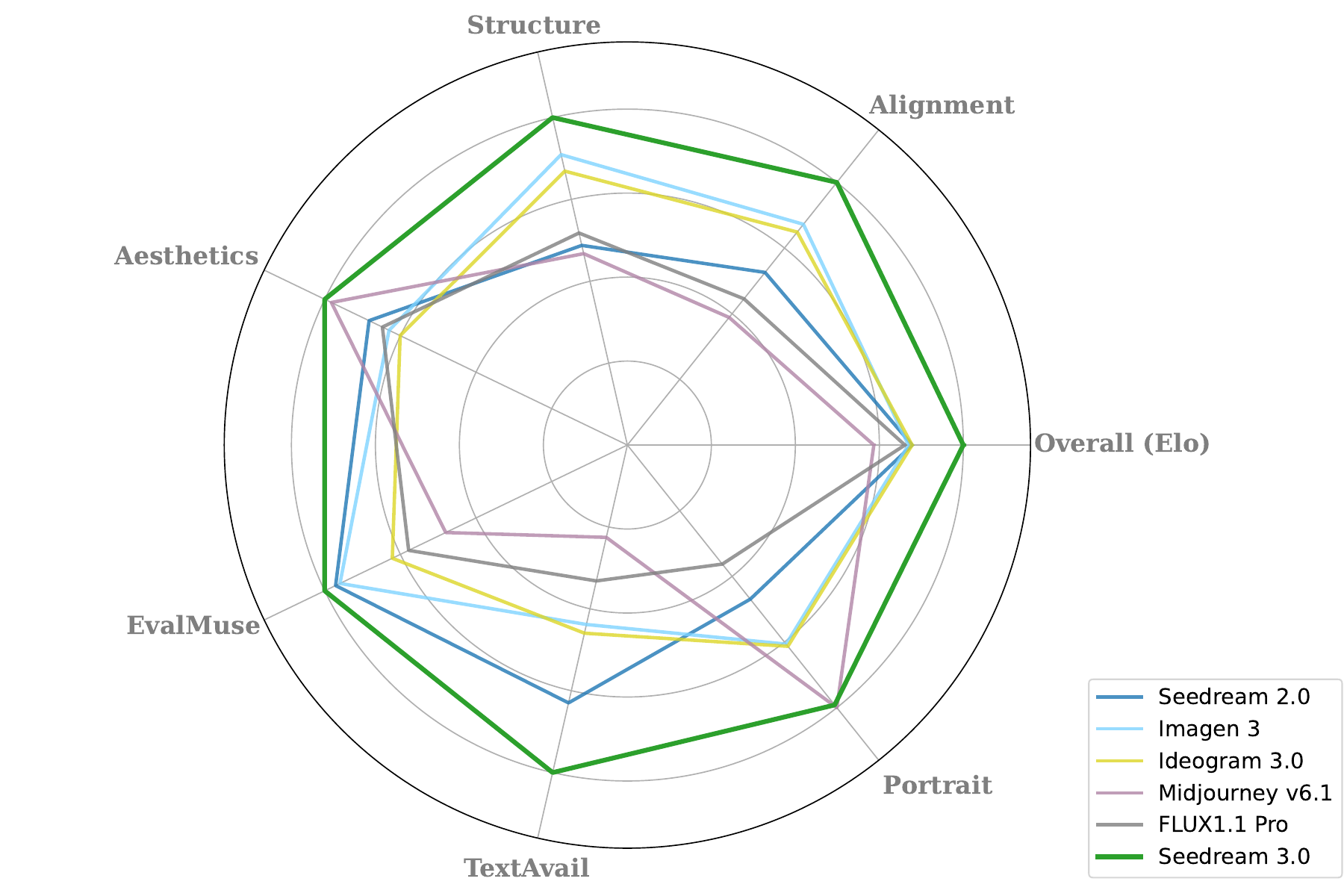}
\end{center}
\label{fig:overall_eval}
\vspace{-5pt}
\caption{Seedream 3.0 demonstrates outstanding performance across all evaluation aspects. Due to missing data, the Portrait result of Imagen 3 and overall result of Seedream 2.0 are represented by the average values of other models. In addition, Seedream 3.0 ranks first at Artificial Analysis Text to Image Model Leaderboard with an Arena ELO score of 1158 at 17.0K
Appearances at the time of publication\protect\footnotemark[1].}
\vspace{-8pt}
\end{figure}
\footnotetext[1]{
\url{https://artificialanalysis.ai/text-to-image/arena?tab=Leaderboard}}

\begin{figure}[pt]
\begin{center}
\includegraphics[width=0.88\linewidth]{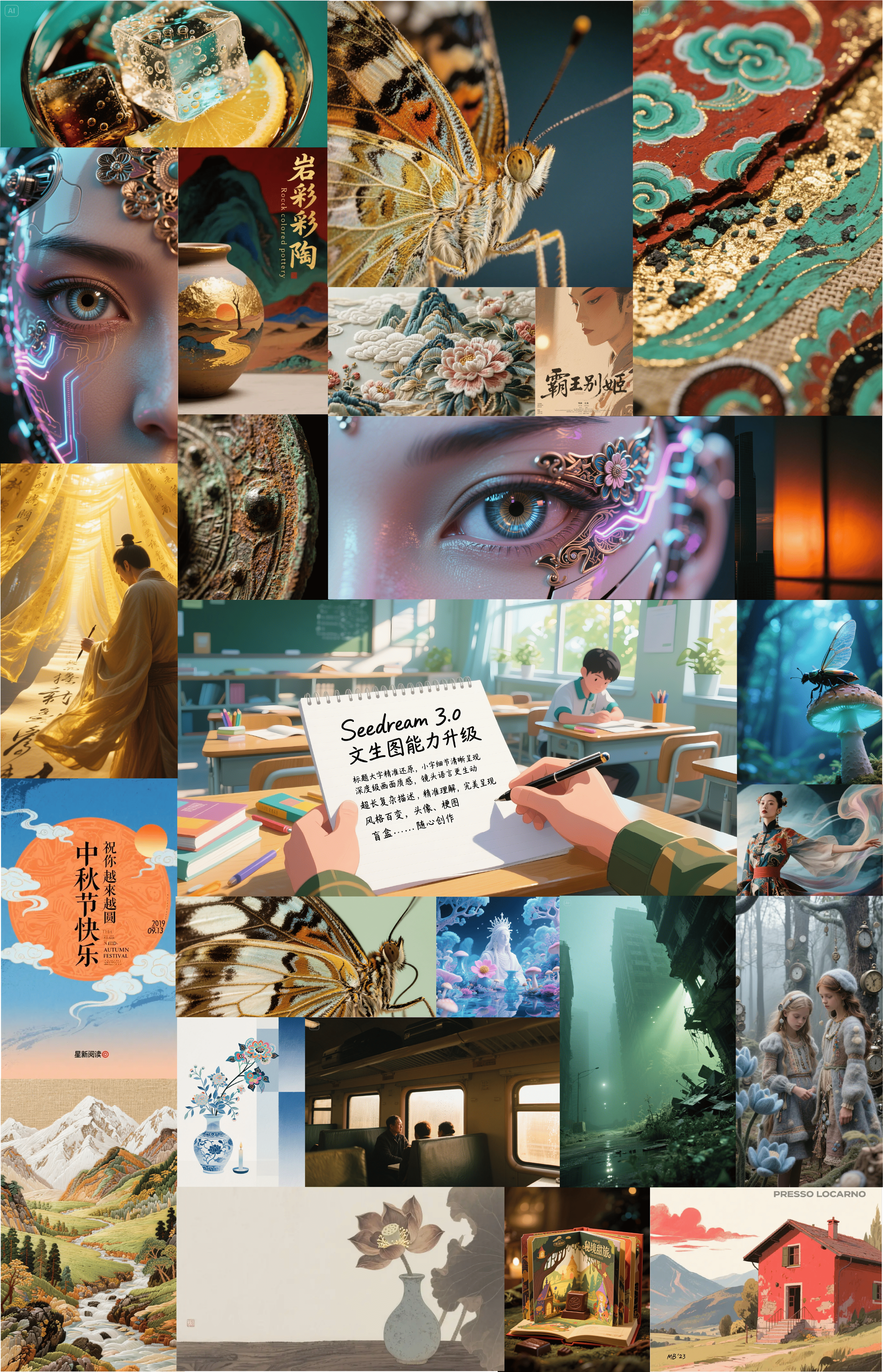}
\end{center}
\label{fig:teaser}
\vspace{-1pt}
\caption{Seedream 3.0 visualization.}
\end{figure}


\clearpage

\tableofcontents

\newpage

\input{sections/Introduction}


\input{sections/technical_details}

\input{sections/Model_Performance}

\clearpage

\bibliographystyle{plainnat}
\bibliography{main}

\clearpage

\beginappendix
\input{sections/appendix}

\end{CJK*}
\end{document}

%% file: resources/packages.tex
\usepackage{natbib}

\usepackage{CJKutf8}

\usepackage{xargs}  

\usepackage{todonotes}  

\usepackage{multirow}

\usepackage{cleveref}

\usepackage{amsmath}
\usepackage{dsfont}

\usepackage{subcaption}

\usepackage{svg}

%% file: sections/Introduction.tex
\section{Introduction}
Recent advances in diffusion models~\cite{ho2020denoising,song2021score,rombach2022high,karras2022elucidating,esser2024scaling} have reshaped the landscape of image generation, propelling generative capabilities to unprecedented heights. Recently, the introduction of Seedream 2.0 has marked a significant milestone in bilingual text-to-image generation, demonstrating superior performance in capturing Chinese linguistic nuances and cultural semantics. However, our comprehensive evaluation identifies several critical challenges that  may impede its wide commercial application.
\begin{itemize}[leftmargin=*]


\item \textbf{\textit{Alignment with complicated prompts}}:
Prompt following can be further enhanced, especially in numerical precision and multi-object spatial relationships. 

\item \textbf{\textit{Fine-grained typographic generation}}:
Seedream 2.0 is still limited in generating high-fidelity small-size text characters, multi-line contextual compositions, and intricate typographic details.

\item \textbf{\textit{Suboptimal visual aesthetics and fidelity}}:
Capturing nuanced aesthetic qualities, such as the beauty of cinematic scenes and the texture of portraits, remains challenging.

\item \textbf{\textit{Limited image resolutions}}:
Fundamental models restrict native output to small resolution (e.g.,512 $\times$ 512px), necessitating reliance on post-processing super-resolution pipelines.

\end{itemize}

Our methodology introduces four key technical improvements. First, at the data stratum, we approximately doubled the dataset size with improved quality by using a new dynamic sampling mechanism, which is built on two orthogonal axes: image cluster distribution and textual semantic coherence.
Second, we incorporate a number of efficient training approaches in the pre-training stage, including i) mixed-resolution training, ii) a cross-modality RoPE, iii) a representation alignment loss, iv) resolution-aware timestep sampling. This allows for better scalability and generalizability, resulting in better visual-language alignment.
Third, in post-training, we utilize diverse aesthetic captions in SFT, and a VLM-based reward model to further enhance the model's overall performance. Finally, in model acceleration, we encourage stable sampling via consistent noise expectation, effectively reducing the number of function evaluations (NFE) during inference.  

Compared to Seedream 2.0, Seedream 3.0 shows significant advances in multiple dimensions:

\begin{itemize}[leftmargin=*]

\item \textbf{\textit{Comprehensive capability enhancement}}：Demonstrates strong user preference and significant advancements in key capabilities, including text-image alignment, compositional structure, aesthetic quality and text rendering.

\item \textbf{\textit{Enhanced text rendering performance}}: Achieves significantly enhanced text rendering performance, particularly excelling in generating small-size text characters in both Chinese and English, and high-aesthetic long-text layouts. Seedream 3.0 represents a pioneering solution for the challenges of small-text generation and aesthetically pleasing long-text composition, outperforming human-designed templates from platforms like Canva in graphic design output.

\item \textbf{\textit{Aesthetic improvement}}: 
Substantial improvement in image aesthetic quality, delivering exceptional performance in cinematic scenarios and enhanced realism in portrait generation.

\item \textbf{\textit{Native high-resolution output}}: Offers native support for 2K resolution output, eliminating the need for post-processing. Also, compatible with higher resolutions and adaptable to diverse aspect ratios.

\item \textbf{\textit{Efficient inference cost}}: With several model acceleration techniques, Seedream 3.0 can reduce its inference cost considerably and generates an image of 1K resolution using only ~{}3.0 seconds (without PE), which is much faster than other commercial models.

\end{itemize}
Seedream 3.0 was integrated into multiple platforms in early April 2025,  including Doubao\protect\footnotemark[1] and Jimeng\protect\footnotemark[2]. We fervently hope that Seedream 3.0 can become a practical tool to improve productivity in all aspects of work and daily life.

\footnotetext[1]{https://www.doubao.com/chat/create-image}
\footnotetext[2]{https://jimeng.jianying.com/ai-tool/image/generate}

%% file: sections/technical_details.tex
\section{Technical Details}

\subsection{Data}
In Seedream 2.0, we employ a stringent data filtering strategy that systematically excluded image data exhibiting minor artifacts, including watermarks, overlaid text, subtitles, and mosaic patterns. This strict filtering protocol significantly limited the amount of data used in the training, especially considering that such affected samples constituted a substantial portion of the original dataset (approximately 35\% of the total collection). To address this limitation, Seedream 3.0 introduces an innovative defect-aware training paradigm. This paradigm includes a specialized defect detector trained on 15,000 manually annotated samples selected by an active learning engine. The detector precisely locates defect areas through bounding box predictions. When the total area of the detected defects is less than 20\% of the image space (a configurable threshold), we retain these previously excluded samples while implementing mask latent space optimization. Specifically, during the diffusion loss calculation in the latent representation space, we employ a spatial attention mask mechanism to exclude feature gradients from the identified defect areas. This innovative approach expands the effective training dataset by 21.7\% while maintaining model stability.

To optimize data distribution, we propose a dual-axis collaborative data sampling framework, jointly optimizing from the dimensions of visual morphology and semantic distribution. In the visual modality, we continue to use hierarchical clustering methods to ensure a balanced representation of different visual patterns. On the textual semantic level, we achieve semantic balance through term frequency and inverse document frequency (TF-IDF~\cite{salton1988term}), effectively addressing the long-tail distribution problem of descriptive texts. To further enhance the coordination of the data ecosystem, we have developed a cross-modal retrieval system that establishes a joint embedding space for image-text pairs. This system achieves state-of-the-art performance across all benchmark tests. The retrieval-enhanced framework dynamically optimizes the dataset through the following methods: (1) injecting expert knowledge via targeted concept retrieval; (2) performing distribution calibration through similarity-weighted sampling; (3) utilizing retrieved neighboring pairs for cross-modal enhancement.

\subsection{Model Pre-training}

\subsubsection{Model Architectures}
Our core architecture design inherits from Seedream 2.0~\cite{gong2025seedream}, which adopts an MMDiT~\cite{esser2024scaling} to process the image and text tokens and capture the relationship between the two modalities. We have increased the total parameters in our base model, and introduced several improvements in Seedream 3.0, leading to enhanced scalability, generalizability, and visual-language alignment.

\textbf{Mixed-resolution Training.} Transformers~\cite{vaswani2017attention} natively supports variable lengths of tokens as input, which also proved to be effective in ViT-based visual recognition tasks~\cite{dehghani2023patch}. In Seedream 3.0, we adopt mixed-resolution training by packing images of different aspect ratios and resolutions together at each training stage. Specifically, we first pre-train our model at an average resolution of $256^2$ (with various aspect ratios) and then finetune it on higher resolution images (from $512^2$ to $2048^2$). We also adopt size embedding as an additional condition to make the model aware of the target resolution. Mixed-resolution training significantly increases data diversity, and improves the generalizability of our model on unseen resolutions.
 
\textbf{Cross-modality RoPE.} In Seedream 2.0, we introduced Scaling RoPE to enable our model to better generalize to untrained aspect ratios and resolutions. In Seedream 3.0, we extend this technique to a Cross-modality RoPE, which further enhances the alignment of visual-text tokens. We treat the text tokens as 2D tokens with the shape of $[1, L]$ and apply a 2D RoPE~\cite{su2024roformer} to the text tokens. The column-wise position IDs of text tokens are assigned consecutively after the corresponding image tokens. The Cross-modality RoPE effectively models the intra-modality and cross-modality relationship, which are crucial for improving visual-text alignment and text rendering accuracy.

\subsubsection{Model Training Details}
\textbf{Training Objectives.} In Seedream 3.0, we adopt flow matching~\cite{lipman2022flow,ma2024sit} training objective, as well as a representation alignment loss (REPA~\cite{yu2024representation}):
\begin{equation}
    \mathcal{L} = \mathbb{E}_{(\mathbf{x}_0, \mathcal{C})\sim \mathcal{D}, t\sim p(t; \mathcal{D}), \mathbf{x}_t\sim p_t(\mathbf{x}_t|\mathbf{x}_0)} \left\|\mathbf{v}_\theta (\mathbf{x}_t, t; \mathcal{C}) - \frac{\dif \mathbf{x}_t}{\dif t}\right\|_2^2 + \lambda \mathcal{L}_{\rm REPA},\label{equ:training_loss}
 \end{equation}
where we use linear interpolant $\mathbf{x}_t = (1 - t)\mathbf{x}_0 + t\bm{\epsilon}, \bm{\epsilon}\sim \mathcal{N}(\bm{0}, \bm{I}) $ following common practice~\cite{esser2024scaling,ma2024sit}. The representation alignment loss is computed as the cosine distance between the intermediate feature of our MMDiT and the feature of a pre-trained vision encoder DINOv2-L~\cite{oquab2023dinov2}, with the loss weight $\lambda=0.5$. We find that introducing the representation alignment objective can accelerate the convergence of large-scale text-to-image generation.

\begin{figure*}[t]
\centering
\includegraphics[width=\linewidth]{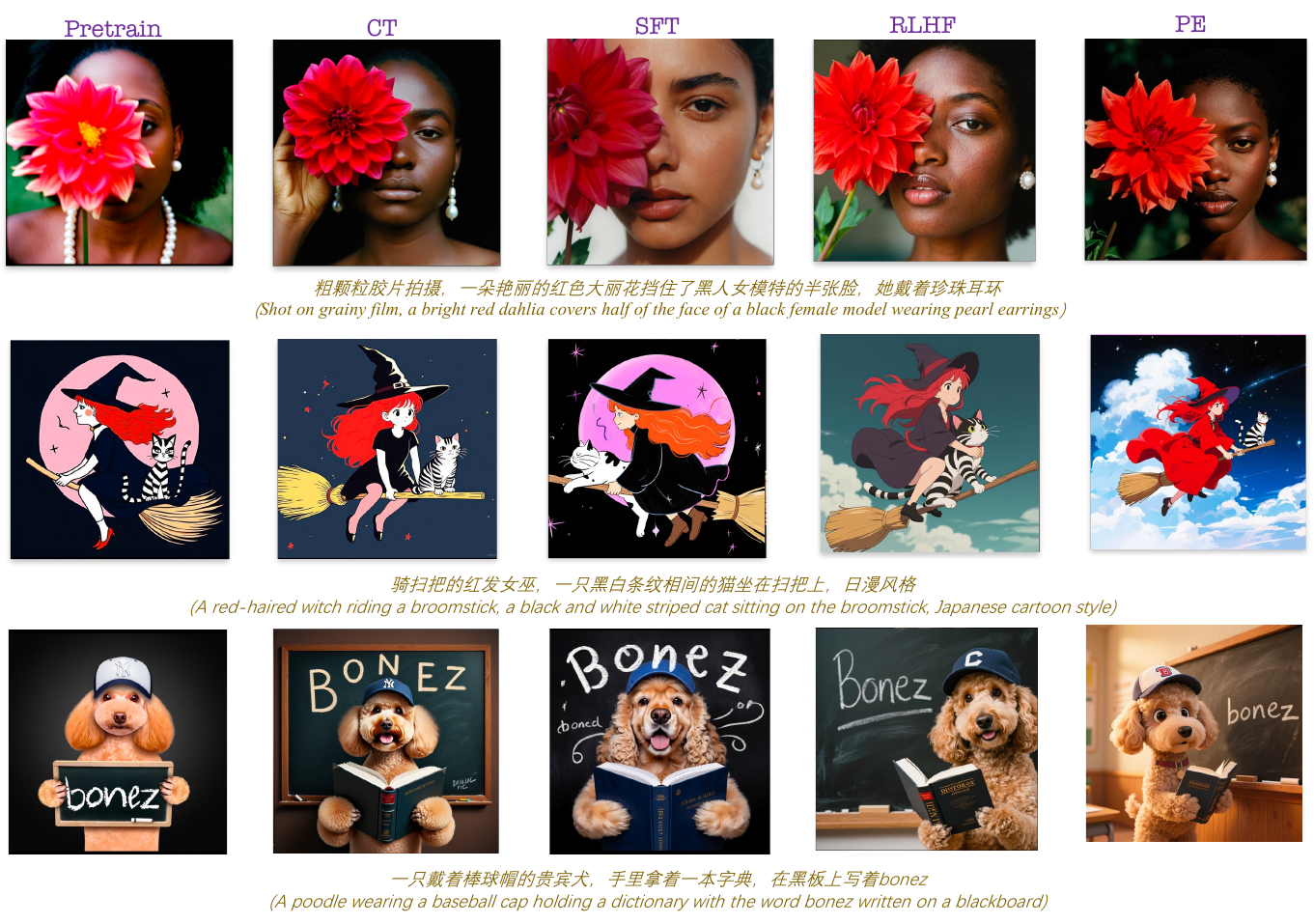}
\caption{The comparison of the effects at different stages.}
\label{fig:post_train}
\end{figure*}

\textbf{Resolution-aware Timestep Sampling.} As shown in Equation~\eqref{equ:training_loss}, the timesteps are sampled from a distribution $p(t; \mathcal{D})$ that is adaptive to dataset $\mathcal{D}$. Similar to~\cite{esser2024scaling}, we design the distribution of timesteps by first sampling from the logit-normal distribution, and then performing timestep shifting based on the training resolution. Generally speaking, when training on higher resolutions, we shift the distribution to increase sampling probability at lower SNRs. During training, we compute the average resolution of dataset $\mathcal{D}$ to determine the shifted timestep distribution. During inference, we compute the shift factor based on the desired resolution and aspect ratio.

\subsection{Model Post-training}
Similar to Seedream 2.0 \cite{gong2025seedream}, our post-training process consists of the following stages: Continuing Training (CT) , Supervised Fine-Tuning (SFT) , Human Feedback Alignment (RLHF) and Prompt Engineering (PE). We omitted the Refiner stage, because our model is capable of directly generating images at any resolution within the range from $512^2$ to $2048^2$. The comparison of the effects at different stages is shown in Figure \ref{fig:post_train}.

\subsubsection{Aesthetic Caption}
\begin{figure*}[t]
\centering
\includegraphics[width=\linewidth]{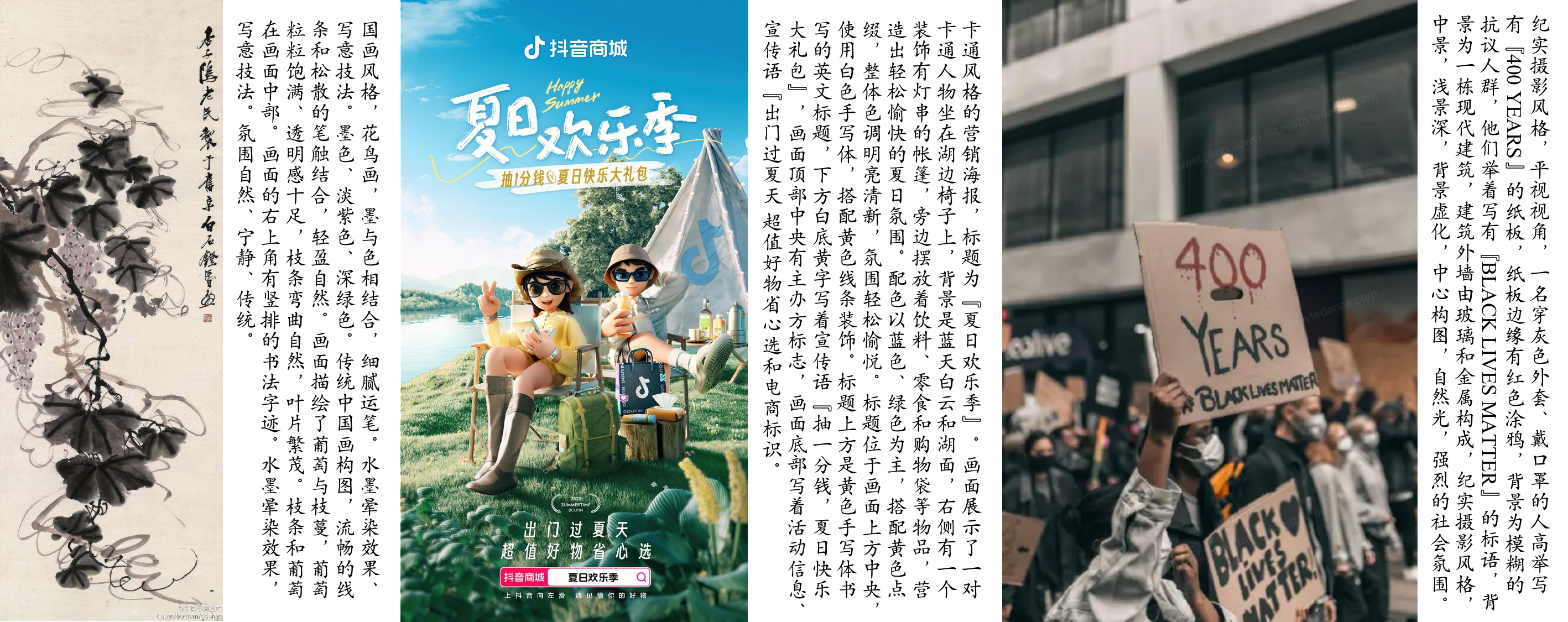}
\caption{Some examples of detailed captions that incorporate aesthetic terms.}
\label{fig:aes_caption}
\end{figure*}

We have specifically trained multiple versions of the caption models for the data in the CT and SFT stages. As shown in Figure \ref{fig:aes_caption}, these caption models provide accurate descriptions in professional domains such as aesthetics, style, and layout. This ensures that the model can respond more effectively to relevant prompts, thereby improving the model's controllability and its performance after prompt engineering.

\subsubsection{Model Training Details}
To ensure that the model could achieve favorable performance across different resolutions, we apply a resolution balancing strategy to the data during the training process. This approach guaranteed an adequate sampling of training data at different resolutions, thereby enhancing the model's ability to follow prompts in various scenarios. 

\subsubsection{Reward Model Scaling}
Different from our previous Seedream 2.0, which employed CLIP as the reward model, we now utilize Vision-Language Models (VLMs) as the reward modeling framework. This change leverages VLMs' superior foundational capabilities and reward scaling potential. Inspired by generative reward modeling (RM) techniques in large language models (LLMs), we explicitly formulate instructions as queries and derive rewards from the normalized probability of the ``Yes'' response token. This approach effectively harnesses the knowledge embedded in preetrained LLMs while naturally benefiting from LLM scaling effects to enhance reward quality. We systematically scale the reward model from 1B to >20B parameters. Empirical results reveal the emergence of reward model scaling, indicating that increased reward model capacity correlates with improved reward modeling performance.

\subsection{Model Acceleration}

Our acceleration framework builds upon Hyper-SD~\cite{ren2025hyper} and RayFlow~\cite{shao2025rayflow}.
We rethink the diffusion process by enabling each sample to follow its own adaptive generative trajectory, rather than forcing all samples through a shared path that converges to a standard Gaussian prior. In conventional diffusion models, all samples are progressively transformed into isotropic Gaussian noise, resulting in overlapping trajectories in probability space. This overlap increases randomness, reduces controllability, and introduces instability during the reverse process. Instead, we guide each data point toward an instance-specific target distribution, enabling trajectory customization per sample. This significantly reduces path collisions and improves both generation stability and sample diversity.

\textbf{Consistent Noise Expectation for Stable Sampling.}  
To ensure smooth and consistent transitions during sampling, we introduce a unified noise expectation vector, estimated from a pretrained model. This expectation serves as a global reference for all timesteps, aligning the denoising process across time. By maintaining consistent expectations, we make it possible to compress the number of sampling steps without degrading image quality. Theoretical analysis further shows that our design maximizes the probability of the forward-backward path from data to noise and back, which leads to improved sampling stability and more reliable reconstructions.

\textbf{Learning to Sample Important Timesteps.}  
In addition to redesigning the generative path, we focus on improving training efficiency. Standard training procedures for diffusion models sample timesteps uniformly, which introduces high variance in the loss and wastes computation on uninformative steps. To address this, we introduce an importance sampling mechanism that learns to focus on the most critical timesteps during training. We achieve this by combining Stochastic Stein Discrepancy~\cite{gorham2020stochastic} (SSD) with a neural network that learns a data-dependent distribution over timesteps. This network predicts which time indices contribute most to reducing the training loss, allowing us to prioritize them during optimization. The result is faster convergence and more efficient use of training resources.


Our framework supports efficient few-step sampling without compromising generation quality. It follows an iterative denoising schedule with far fewer steps than unaccelerated baselines. Despite this reduction, our method achieves results that match or surpass baselines requiring 50 function evaluations—known as the Number of Function Evaluations (NFE)—across key aspects including aesthetic quality, text-image alignment, and structural fidelity. These results demonstrate the effectiveness of our trajectory design and noise consistency mechanisms in enabling high-quality synthesis with minimal computational cost.
For other acceleration methods, such as Quantization, we directly follow the solution of Seedream 2.0.

%% file: sections/Model_Performance.tex
\section{Model Performance}

In a publicly conducted evaluation, Seedream 3.0 ranks first  among top-tier text-to-image models globally, such as GPT-4o~\cite{gpt-4o}, Imagen 3~\cite{imagen3}, Midjourney v6.1~\cite{mjv6}, FLUX1.1 Pro~\cite{flux2023}, Ideogram 3.0~\cite{ideogram}, and others.
We further conduct a rigorous expert evaluations to assess Seedream 3.0, both manually and through automated means. The results  demonstrate marked improvements in Seedream 3.0 across all key performance indicators compared to the previous version, alongside superior performance against industry-leading counterparts. Notably, Seedream 3.0 exhibits achieves exceptional capabilities in two aspects: dense text rendering and  photorealistic human portrait generation. See Sections 3.3 and 3.4 for detailed explanations of these two aspects, respectively. In addition, we provide a systematic comparative analysis with GPT-4o~\cite{gpt-4o} in Section 3.5, exploring the capability boundaries of the two models in different fields.
The overall results are presented in Figure 1.

\subsection{Artificial Analysis Arena}

Artificial Analysis~\cite{aa_elo} is a leading benchmarking platform for AI models, specifically focused on image and video generation. It offers dynamic leaderboards that evaluate models based on key metrics such as output quality, generation speed, and cost, providing an objective comparison of state-of-the-art AI systems. The Text-to-Image leaderboard allows users to anonymously compare the generated images from different models. This ensures fairness, as users vote on images generated using identical prompts without knowing what the models are. Models are ranked using an ELO scoring system, which reflects user preferences to some extent. 

Seedream 3.0 participated in the Artificial Analysis ranking and secured the top position overall, outperforming GPT-4o and establishing a substantial lead over other models, including Recraft V3, HiDream, Reve Image, Imagen 3 (v002), FLUX1.1 Pro, and Midjourney v6.1. Additionally, it demonstrates the best performance across most sub-dimensions, including \textit{Style} categories such as General \& Photorealistic, Anime, Cartoon \& Illustration, and Traditional Art, as well as \textit{Subject} categories such as People: Portraits, People: Groups \& Activities, Fantasy, Futuristic, and Physical Spaces. 

\begin{figure}[t]
    \centering
    \includegraphics[width=\linewidth]{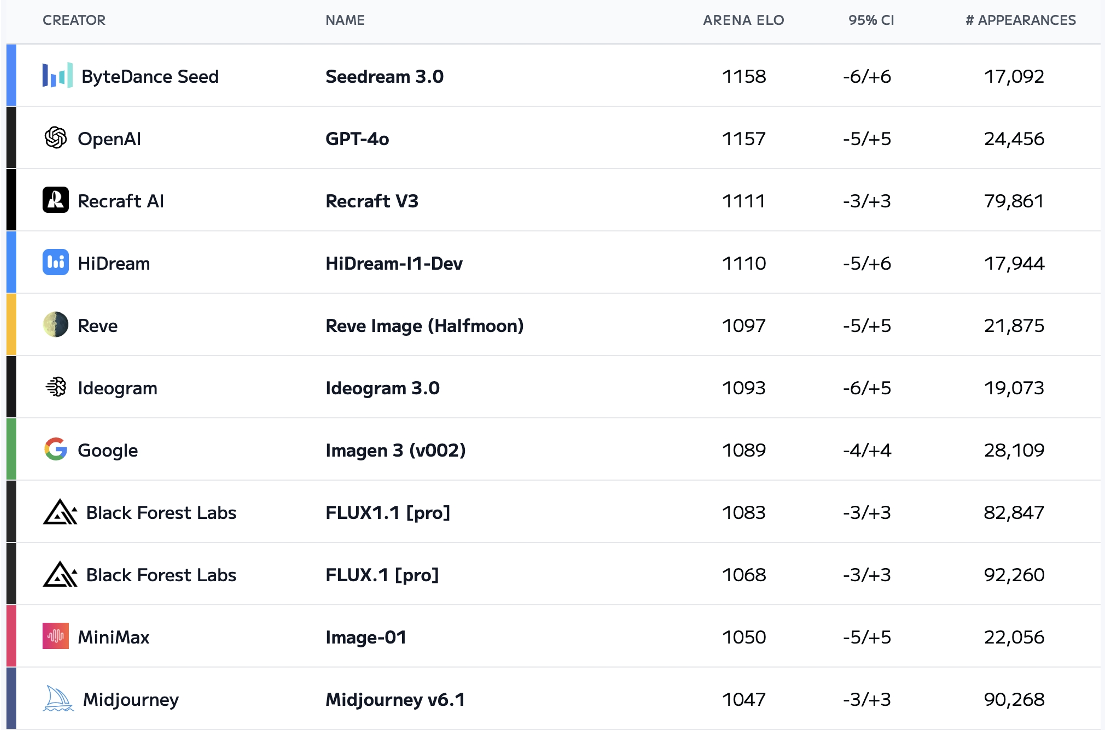}
    \caption{Results from Artificial Analysis Arena.}
    \label{fig:aa_seedream3.0}
\end{figure}

\begin{figure*}[h]
\centering
\includegraphics[width=\linewidth]{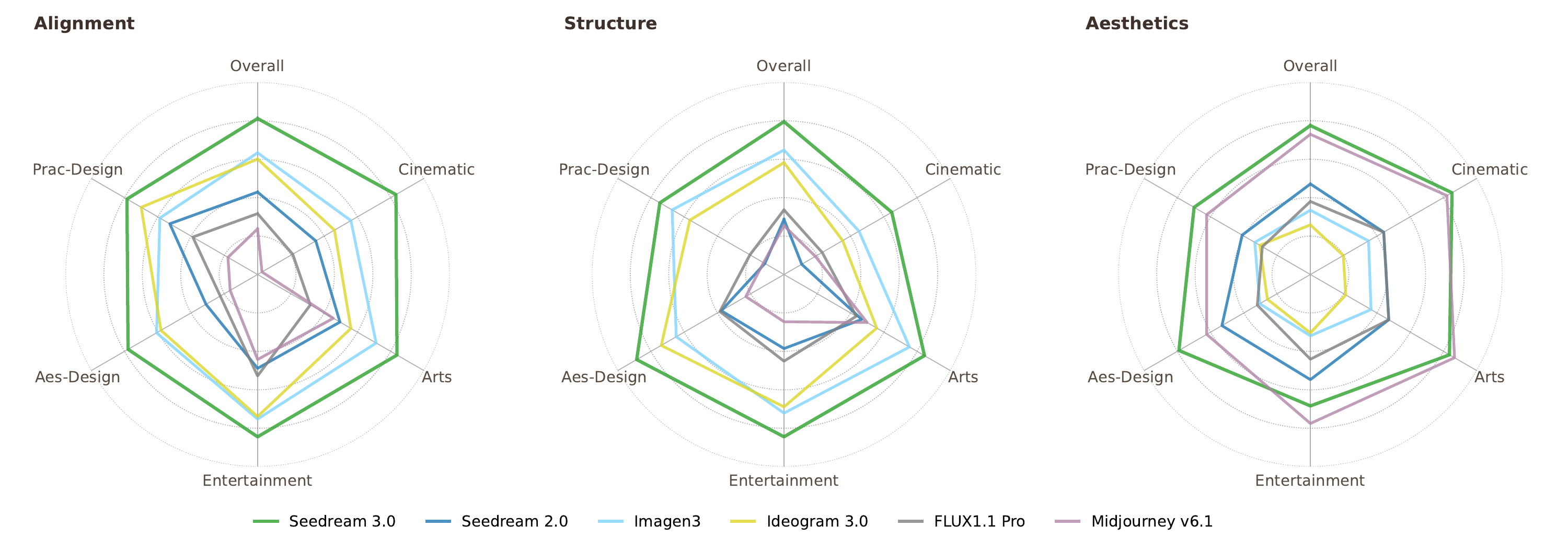}
\caption{Human evaluation results.}
\label{fig:human_eval_res}
\end{figure*}

\begin{figure*}[h]
\centering
\includegraphics[width=\linewidth]
{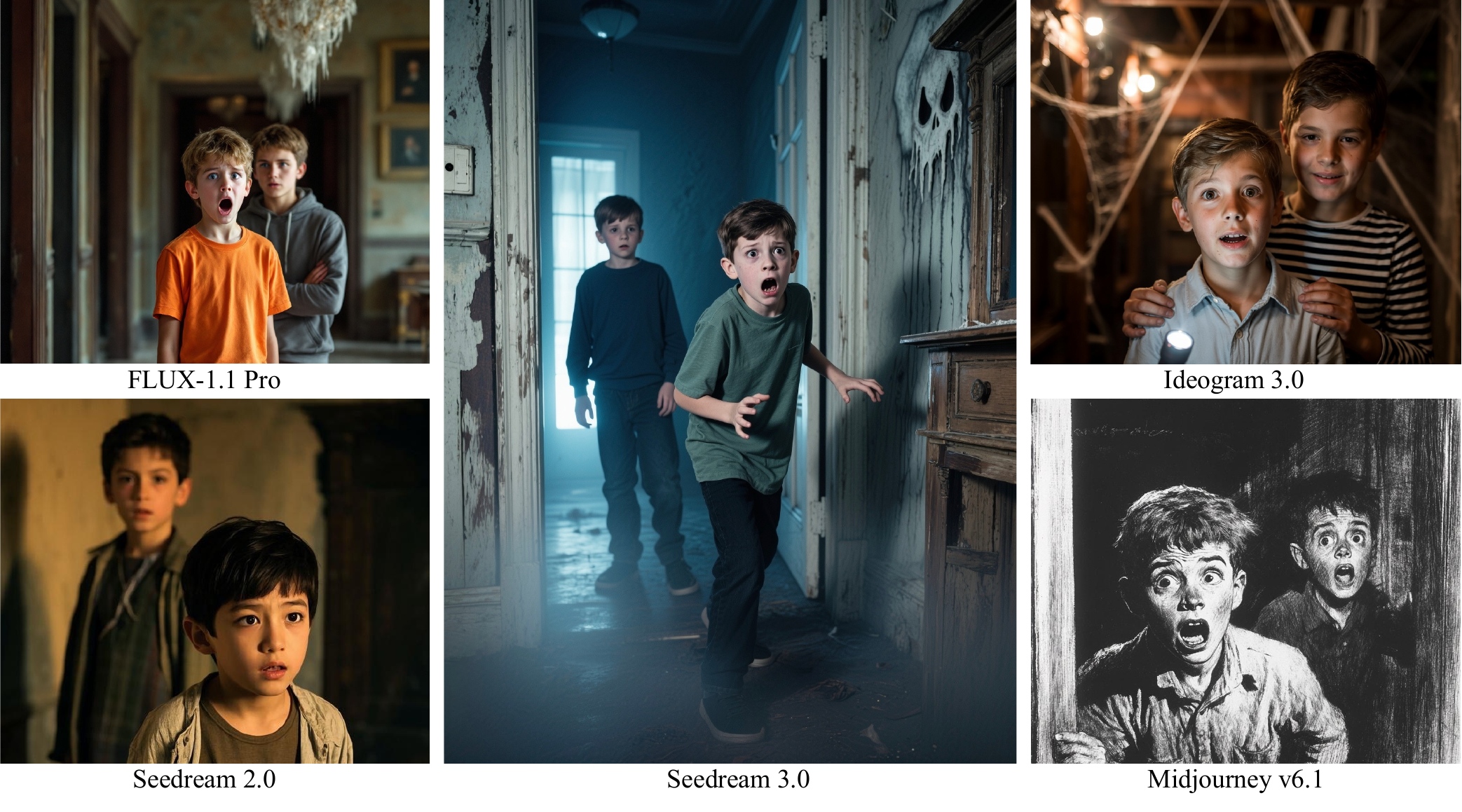}
\caption{Alignment Comparison. Prompt: Two boys are in the haunted house. The boy in the front looks frightened, while the boy behind appears calm.}
\label{fig:human_align}
\end{figure*}

\begin{figure*}[!t]
\centering
\includegraphics[width=\linewidth]{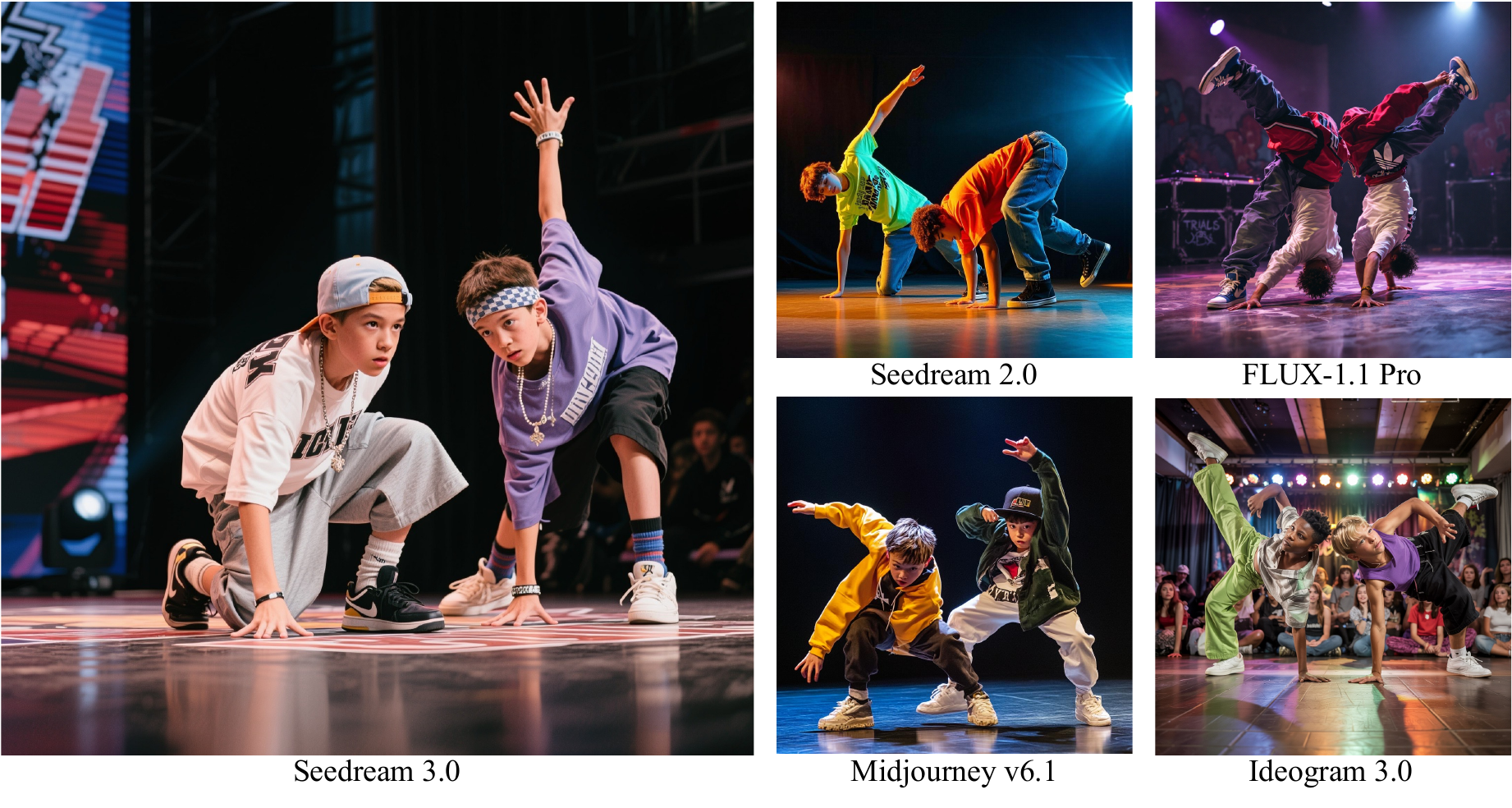}
\caption{Structure Comparison. Prompt: Two 14-year-old boys, dressed in Y2K style, perform a one-handed ground move on stage as part of a breakdancing routine. \textbf{Warning: These images may cause discomfort.}}
\label{fig:human_structure}
\end{figure*}

\subsection{Comprehensive Evaluation}
\subsubsection{Human Evaluation}
A larger evaluation benchmark is established to conduct a more comprehensive evaluation of Seedream 3.0 in different scenarios. This benchmark, named Bench-377, is made up of 377 prompts. 
In addition to examining basic dimensions such as text-to-image alignment, structure plausibility, and aesthetic sense, the design of prompts also takes into account the usage scenarios. We consider five main scenarios: cinematic, arts, entertainment, aesthetic design, and practical design. We propose the practical design category as Seedream 3.0 is proved to be helpful in assisting routine work and studying. For example, it can provide support in tasks such as icon arrangements in slides and illustration design in handwriting newspapers.

\begin{figure*}[h]
\centering
\includegraphics[width=\linewidth]{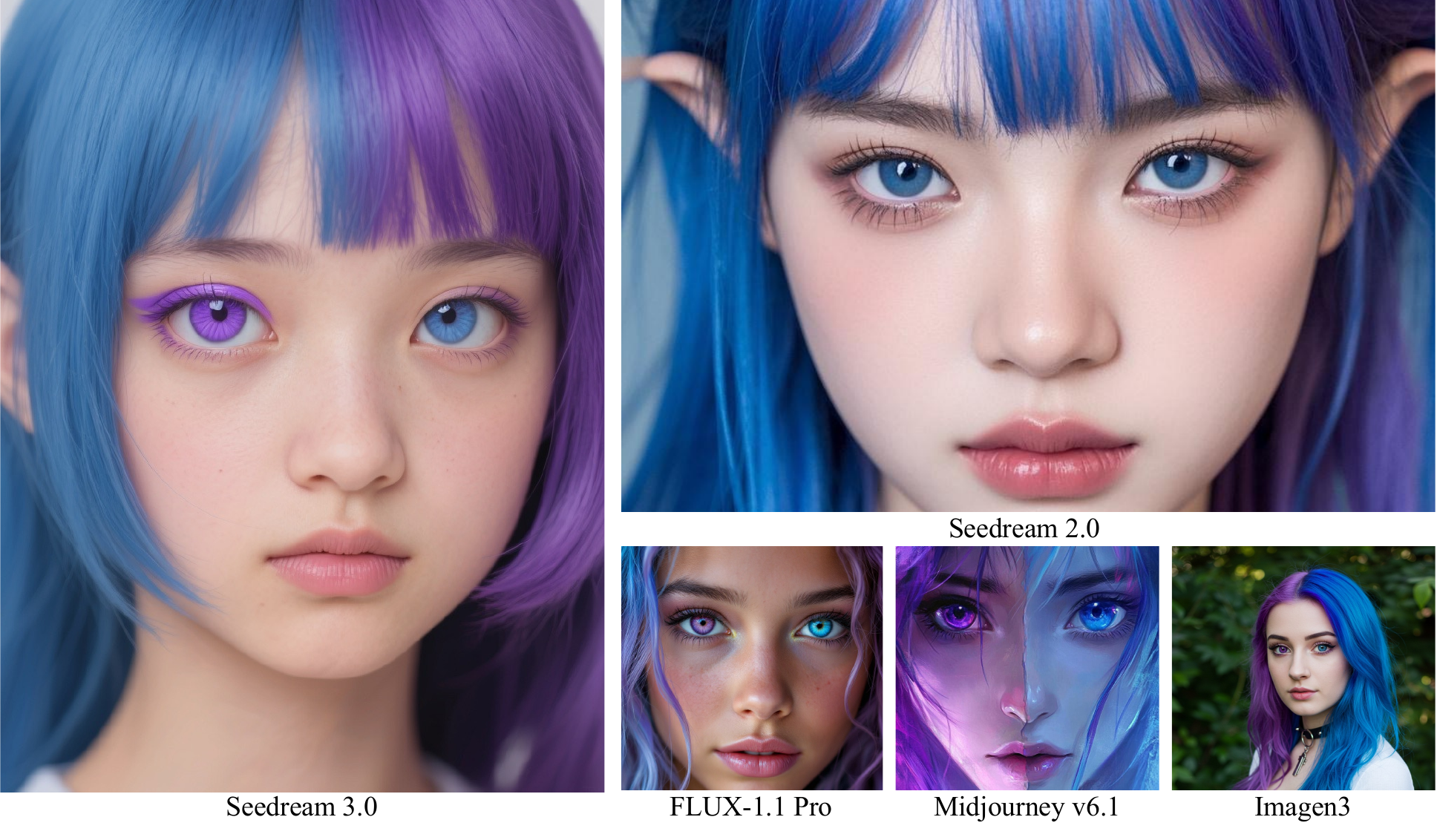}
\caption{Aesthetic Comparison. Prompt: A girl, one eye is purple, and the hair on that side is blue. The other eye is blue, and the hair on that side is purple. realistic.}
\label{fig:human_aes}
\end{figure*}

\begin{figure*}[!t]
\centering
\includegraphics[width=\linewidth]{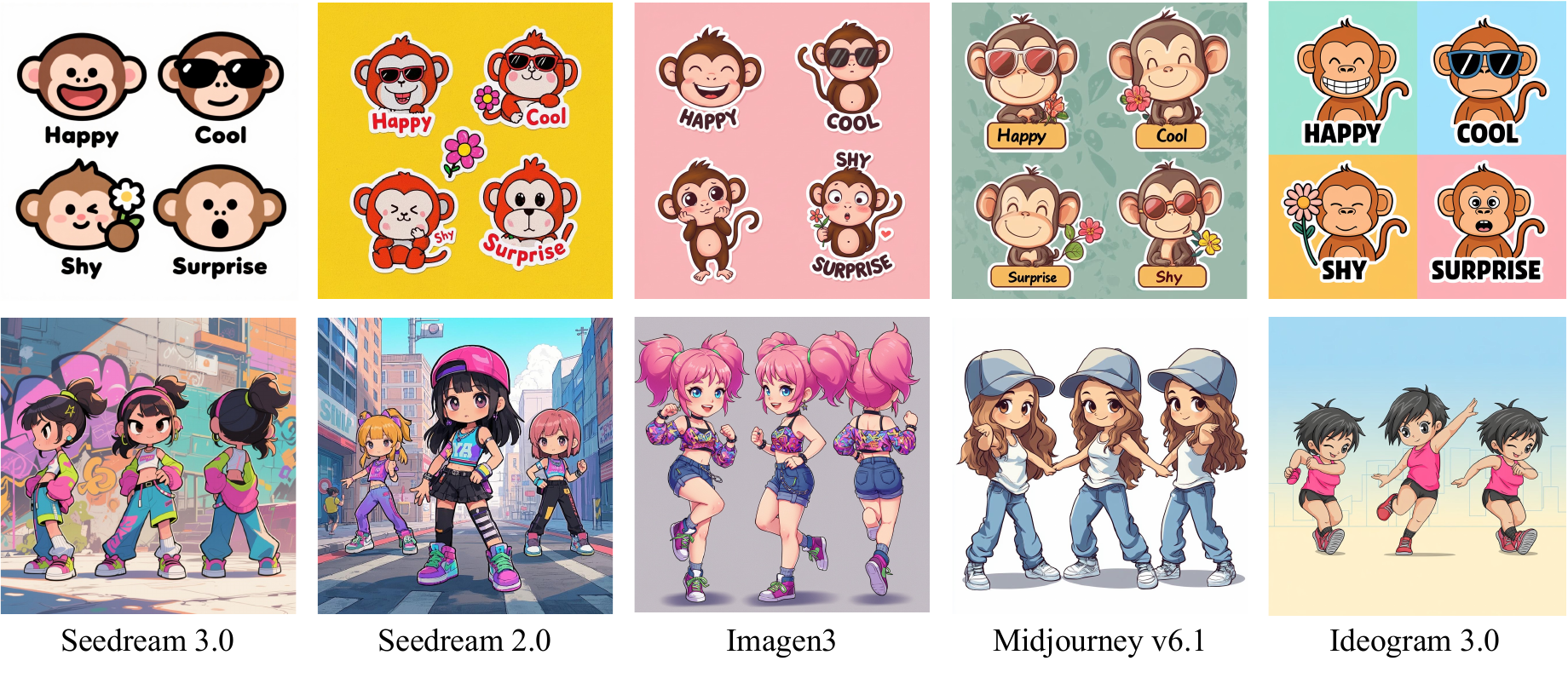}
\caption{Design Comparison. \textbf{Top Prompt}: Sticker Series Design: Sticker 1: A monkey is grinning with the text ``Happy'' below. Sticker 2: The monkey wears sunglasses with the text ``Cool'' below. Sticker 3: The monkey is holding a flower with a shy expression, with the text ``Shy'' below. Sticker 4: The monkey looks surprised, with the text ``Surprise'' below. \textbf{Bottom Prompt}: Chibi character, girl, full body, street dance, three-view drawing.}
\label{fig:design_comp}
\end{figure*}


A systematic evaluation by human experts of text-to-image models was performed based on Bench-377. The evaluation is carried out using three basic criteria: text-image alignment, structural correction, and aesthetic quality. The specific results for the five usage scenarios are presented in Figure~\ref{fig:human_eval_res}. 
Seedream 3.0 significantly outperforms Seedream 2.0 and competing models across text-image alignment and structural fidelity. Notably, it achieves an overall score higher than that of Midjourney in terms of aesthetic performance. Moreover, it is notably superior to it in the design category, though it lags slightly behind in categories such as art. While Imagen 3 also demonstrates competent performance in text-image alignment and structure, it underperforms in aesthetic evaluation. Midjourney exhibits superior aesthetic capabilities but shows limited proficiency in functional alignment and structural fidelity. 

Figures~\ref{fig:human_align},\ref{fig:human_structure},\ref{fig:human_aes}, and \ref{fig:design_comp} illustrate how enhanced fundamental capabilities facilitate the generation of diverse scenarios.
Improved text-to-image alignment enables more precise representation of user intentions.
For example, the lively depiction of micro-expressions improves the portrayal of a movie's atmosphere.
Precise understanding and expression of complex descriptions and specialized terms, such as "three-view", effectively fulfill users’ design requirements. These capabilities are fundamentally supported by enhanced structural stability and aesthetic quality. For example, the integrity of the limbs in dynamic motions, the detailed presentation of small objects, as well as improved capabilities in color, lighting, texture, and composition are all instrumental to the high availability of Seedream 3.0.

\begin{table*}[t]
      \caption{Preference evaluation with different metrics.}
  \centering
\begin{tabular}{p{2.3cm}|p{1.6cm} p{2cm} p{1.6cm} p{1.6cm} p{2.2cm} p{2.2cm}}
    \toprule
    Metirc & FLUX1.1  & Ideogram 2.0 & MJ v6.1 & Imagen 3 & Seedream 2.0 & Seedream 3.0  \\
    \midrule
 EvalMuse & ~~0.617&~~0.632&~~0.583 & ~~0.680 &~~~0.684&~~~\textbf{0.694} \\   
 HPSv2 & ~0.2946&~0.2932&~0.2850&~0.2951 &~~0.2994&~~~\textbf{0.3011} \\
 MPS & ~~13.11&~~13.01&~~13.67&~~13.33&~~~13.61&~~~\textbf{13.93}  \\
 \midrule
 Internal-Align & ~~27.75&~~27.92 &~~28.93  &~~28.75 &~~~29.05 &~~~\textbf{30.16}  \\
Internal-Aes    & ~~25.15&~~26.40 &~~27.07  &~~26.72 &~~~26.97 &~~~\textbf{27.68}  \\

    \bottomrule
    \end{tabular}
  \label{tab:image_quality}%
\end{table*}%

\subsubsection{Automatic Evaluation}

In accordance with the automatic evaluation of the previous version, we assess the text-to-image generation model based on two criteria: text-image alignment and image quality. Seedream 3.0 consistently ranks first across all benchmarks.

For automatic evaluation for text-to-image alignment, we mainly focus on EvalMuse~\cite{han2024evalmuse40kreliablefinegrainedbenchmark}, which exhibits relatively good consistency with human evaluations across multiple benchmarks.
Seedream 3.0 outperforms other models as shown in Table~\ref{tab:image_quality}.
Further analysis in the fine-grand dimension shows that, compared to Seedream 2.0, Seedream 3.0 has improvements in most dimensions, especially in terms of objects, activities, locations, food, and space.
To align with the previous reported results, Ideogram 2.0 is included in the assessment here and subsequent chapters.

For image quality evaluation, we reuse two external metrics, HPSv2~\cite{wu2023human} and MPS~\cite{MPS}, and two internal evaluation models, Internal-Align and Internal-Aes. Seedream 3.0 ranks first in all metrics as shown in Table~\ref{tab:image_quality}.
In the aesthetic evaluation, which includes MPS and our in-house aesthetic evaluation models, Seedream 3.0 outperforms Midjourney, while Seedream 2.0 didn't in previous assessments. At the same time, in terms of the HPSv2 index, Seedream3.0 exceeds 0.3 for the first time, indicating that our model has excellent consistency with human preferences.


\subsection{Text Rendering}
Seedream 2.0's text rendering, particularly for Chinese characters, has garnered widespread acclaim from users. In Seedream 3.0, we have further optimized this capability and conducted thorough evaluations. Our text evaluation benchmark comprises 180 Chinese prompts and 180 English prompts, covering a diverse range of categories, including logo designs, posters, electronic displays, printed text, and handwritten text.

One perception-based metric, \textbf{availability rate}, and two statistics-based metrics, text \textbf{accuracy rate} and \textbf{hit rate}, are employed to evaluate text rendering capability.
The availability rate refers to the proportion of images deemed acceptable when text rendering is generally correct, taking into account the integration of text with other content and the overall aesthetic quality. The objective metrics are defined as follows:

\begin{figure*}[!t]
\centering
\includegraphics[width=\linewidth]{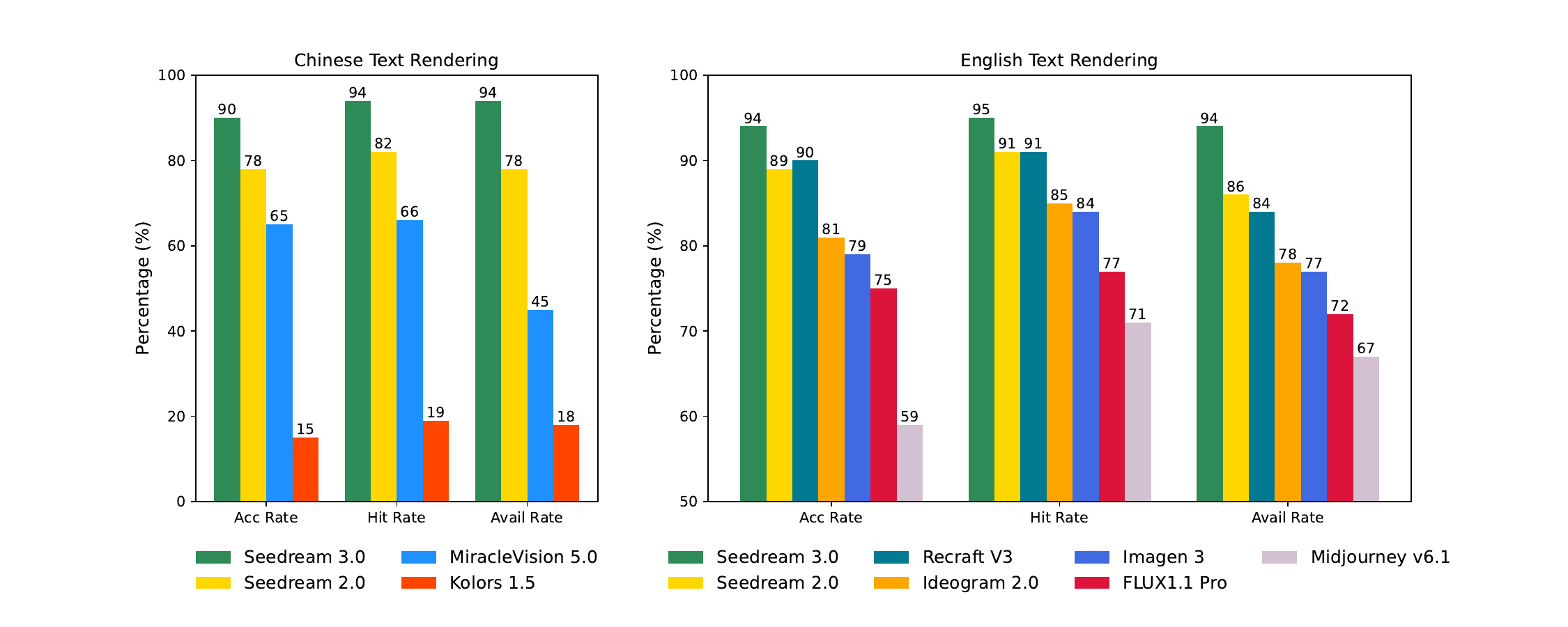}
\caption{Text Rendering Evaluation.}
\label{fig:text_rendering}
\end{figure*}

\begin{figure*}[!t]
\centering
\includegraphics[width=\linewidth]{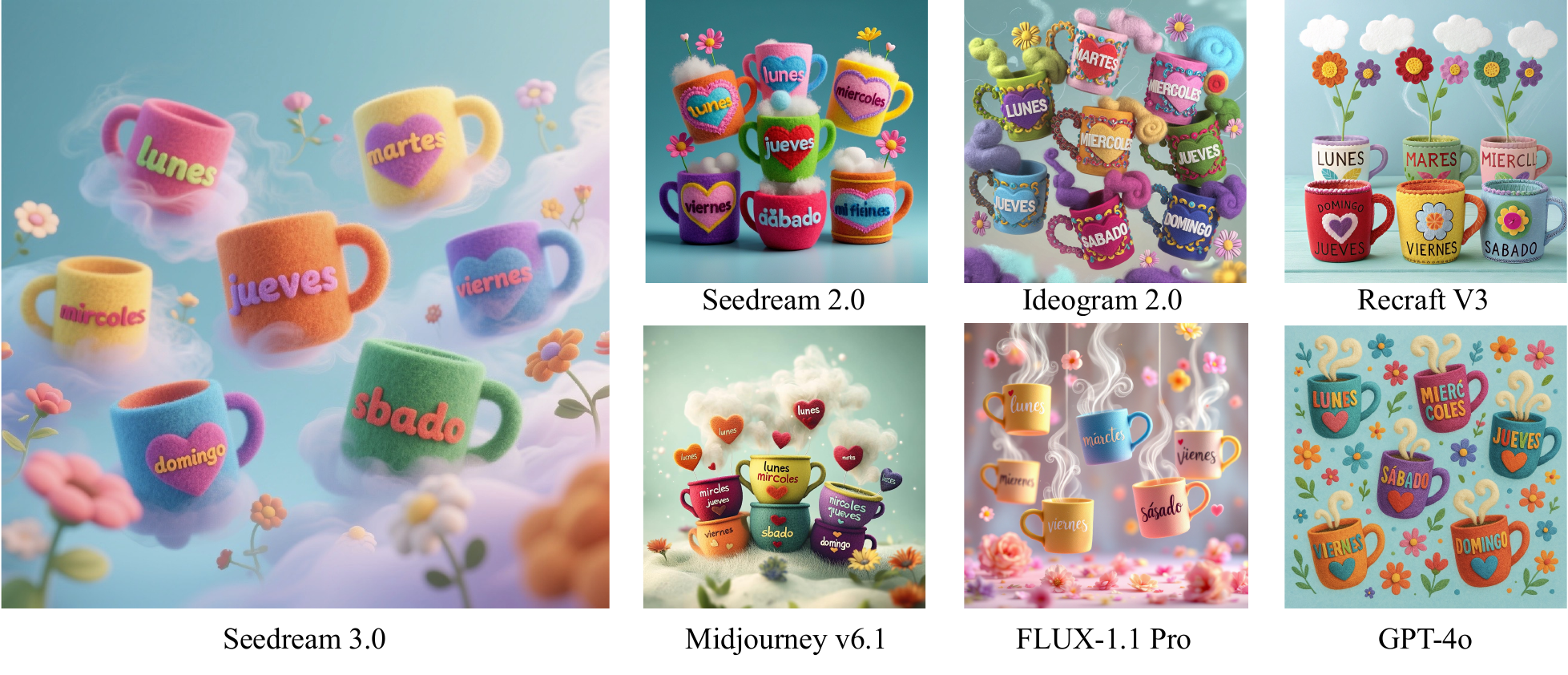}
\caption{Text Rendering comparisons. Prompt: A captivating and vibrant image, 3D render, featuring seven colorful, ornate felt mugs, each adorned with a heart and displaying bold text representing the days of the week: ``lunes'', ``martes'', ``mircoles'', ``jueves'', ``viernes'', ``sbado'', ``domingo''. These lively mugs are filled with whimsical felt smoke, and they elegantly float in a dreamy, enchanting atmosphere. The diverse array of floating flowers adds depth and dimension to the scene, while the soft baby blue background harmoniously complements the design. fashion, illustration, typography, 3d render, painting.}
\label{fig:text_comp}
\end{figure*}

\begin{itemize}
\vspace{-1mm}
    \item \textbf{Text accuracy rate} is defined as:
    \vspace{-1mm}
    \[
    R_a = \left(1 - \frac{N_e}{N}\right) \times  100\%
    \]

    where \(N\) represents the total number of target characters, and \(N_e\) denotes the minimum edit distance between the rendered text and the target text.
    \item \textbf{Text hit rate} is defined as:
    \[
    R_h = \frac{N_c}{N} \times 100\%
    \]
    \vspace{-1mm}
    where \(N_c\) represents the number of characters correctly rendered in the output.
    \vspace{-1mm}
\end{itemize}


\begin{figure*}[!t]
\centering
\includegraphics[width=\linewidth]{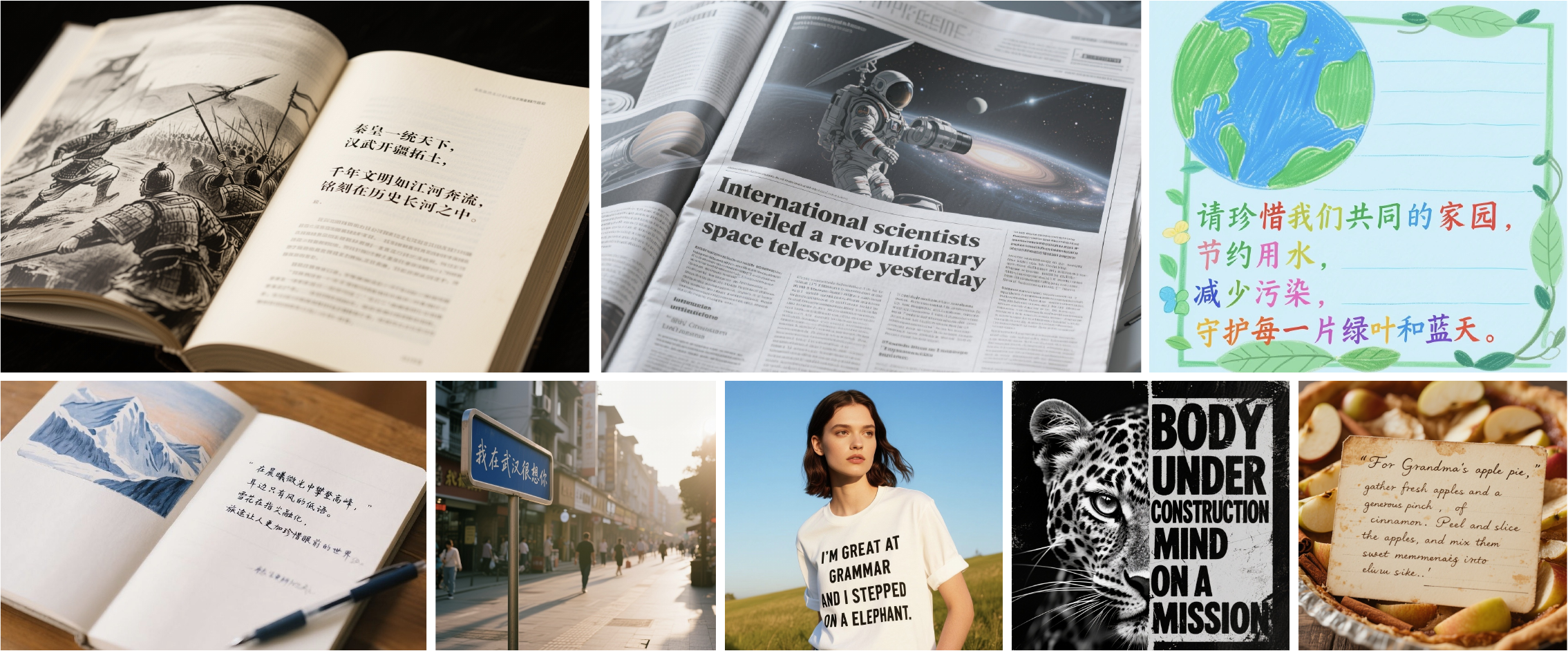}
\caption{Text Rendering by Seedream 3.0.}
\label{fig:text_showcase}
\end{figure*}

Figure~\ref{fig:text_rendering} demonstrates that Seedream 3.0 achieves superior text rendering performance compared to existing models, including its predecessor (Seedream 2.0). The system achieves a 94\% text availability rate for both Chinese and English characters, effectively eliminating text rendering as a limiting factor in image generation. Notably, Chinese text availability shows an improvement of 16\% over Seedream 2.0. The nearly equivalent values of availability and hit rates further indicate minimal occurrence of layout or medium-related rendering errors. These results validate the effectiveness of our native text rendering approach compared to post-processing composition methods and external plugin solutions.

In addition to the overall improvement in availability rate, it is crucial to highlight the exceptional performance of Seedream 3.0 in rendering dense text. Dense text, characterized by long passages with a high density of small characters, such as greetings with numerous words, has posed a challenge for previous models. In contrast, Seedream 3.0 shows significant advancements in handling such fine characters. As illustrated in Figures \ref{fig:text_comp} and \ref{fig:text_showcase}, Seedream 3.0 excels in both the precision of small character generation and the naturalness of text layout. For comparison, GPT-4o, another model known for its dense text rendering capabilities, will be evaluated in the following sections.

\begin{figure*}[!t]
\centering
\includegraphics[height=7cm]{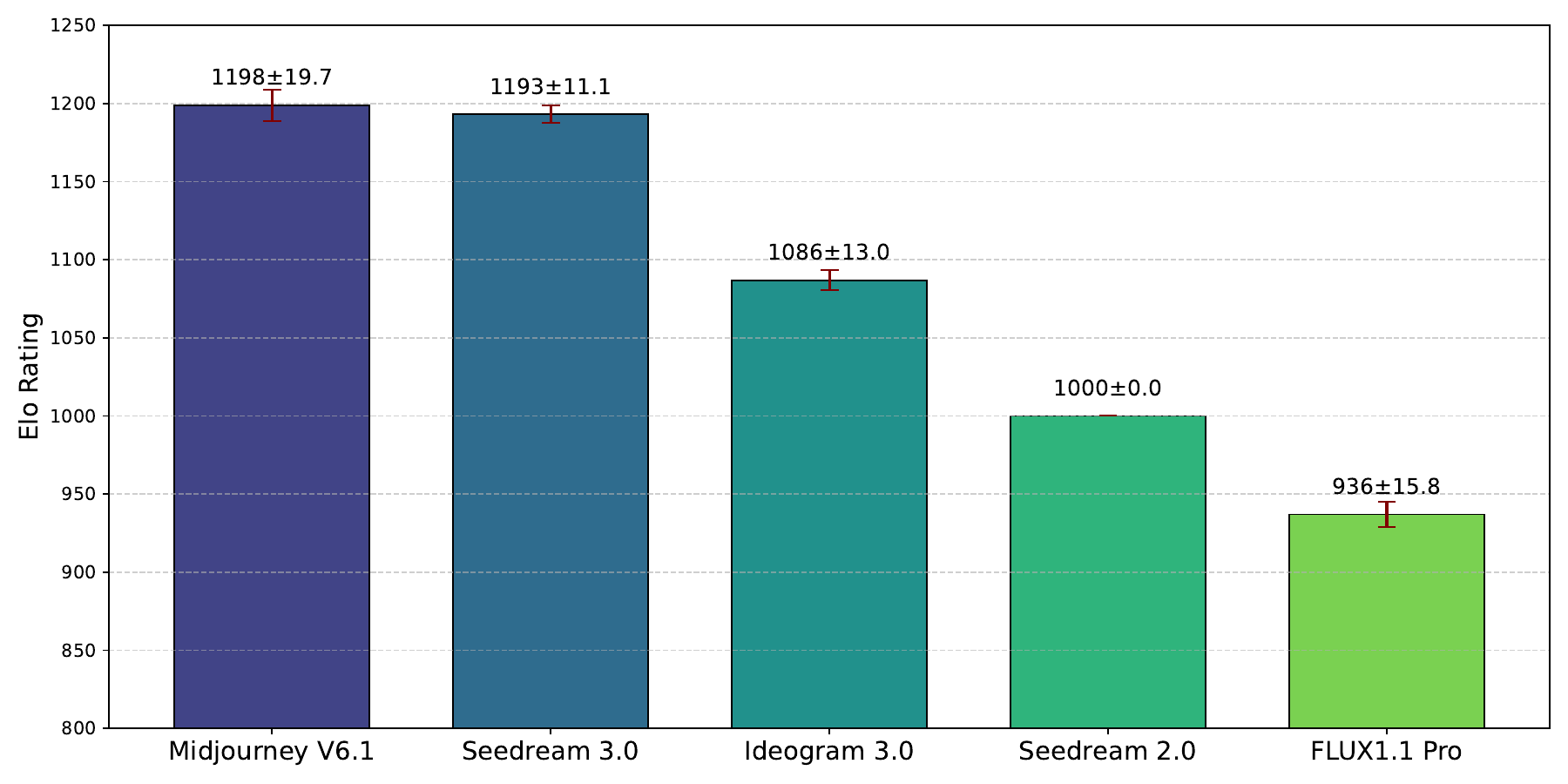}
\caption{Photorealistic Portrait Evaluation.}
\label{fig:portarit_elo}
\end{figure*}

\begin{figure*}[!b]
\centering
\includegraphics[width=\linewidth]{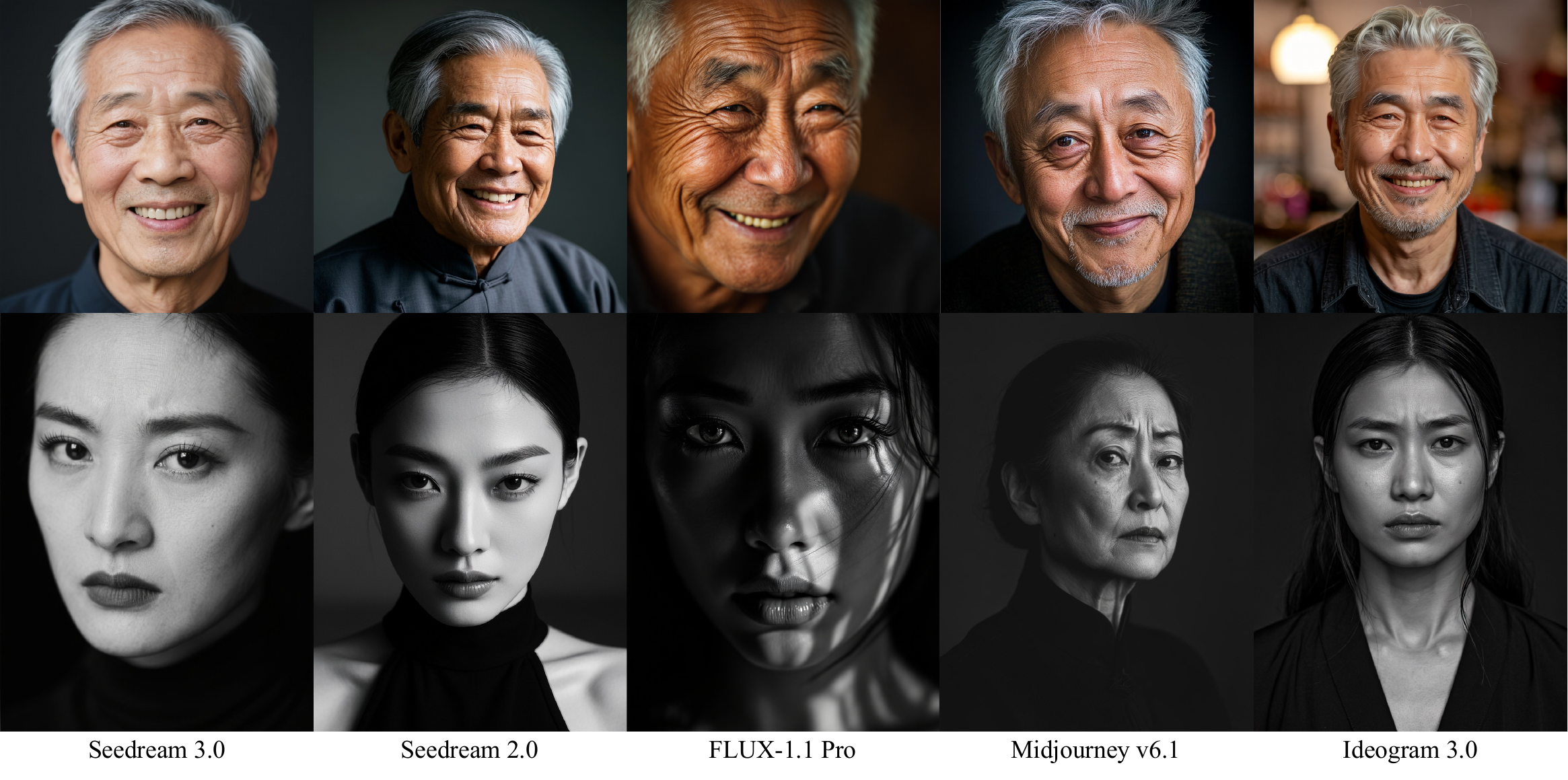}
\caption{Realistic Portrait comparisons.}
\label{fig:human_comp}
\end{figure*}

\begin{figure*}[t]
\centering
\includegraphics[width=\linewidth]{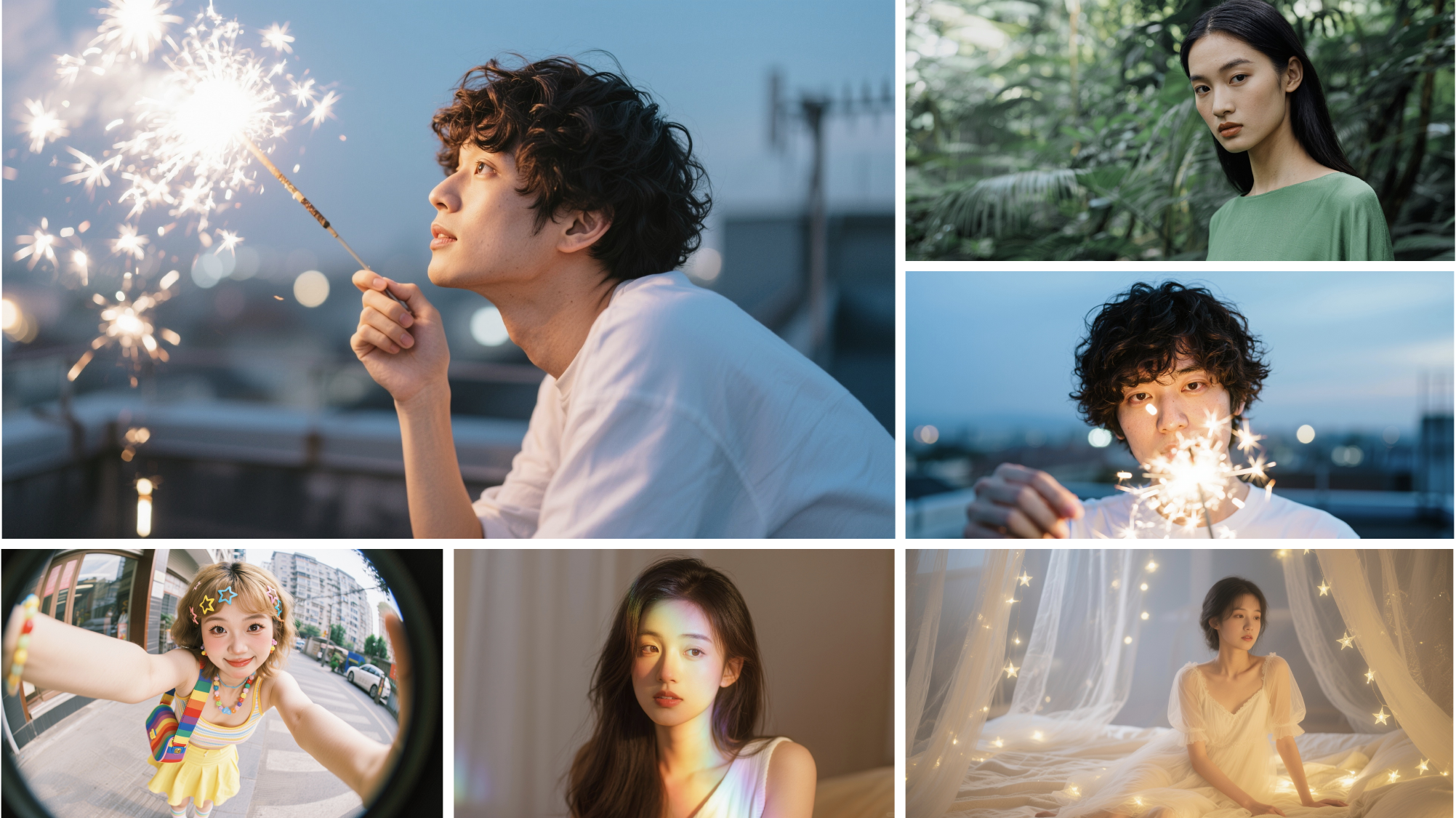}
\caption{Human Portraits from Seedream 3.0 with higher resolution. High resolution provides enhanced texture and clarity.}
\label{fig:human_showcase}
\end{figure*}

\subsection{Photorealistic Portrait}

The overly synthetic appearance of AI-generated images, especially in portraits, has long been a criticism of Text-to-Image models. Issues like overly smooth skin and an oily texture make the generated images appear artificial.

To comprehensively assess Seedream 3.0's performance in this area, we construct a portrait evaluation set comprising 100 prompts. These prompts focus on various aspects of portrait generation, including expressions, postures, angles, hair features, skin texture, clothing, and accessories. The evaluation follows an Elo battle approach, where participants are asked to select their preferred portraits generated by different models and justify their choice. The evaluation criteria focus on two primary dimensions: realism and emotion. Competitors include Seedream 3.0, Seedream 2.0, Midjourney v6.1, FLUX-Pro 1.1, and the recently updated Ideogram 3.0, known for its photorealistic generation.
To ensure a fair comparison, multiple rounds of image generation are performed for Midjourney v6.1 to ensure a realistic result, avoiding those that are overly artistic or abstract.

After a public evaluation involving over 50,000 battle rounds, we obtain the results as shown in Figure~\ref{fig:portarit_elo}. Note that some model variants are not displayed. Seedream 3.0 and Midjourney v6.1 both rank first, significantly outperforming other models. Examples in Figure~\ref{fig:human_comp} demonstrate that Seedream 3.0 effectively eliminates the artificial appearance. In portrait generation, the skin textures now exhibit realistic features such as wrinkles, fine facial hair, and scars, closely resembling natural human skin. Meanwhile, Seedream 3.0 can still generate flawless skin textures when prompted. Additionally, while the texture of portraits generated by Midjourney v6.1 appears slightly inferior to Seedream 3.0, it excels in conveying emotional expressions, contributing to its high ranking. Future versions will aim to further enhance both aspects.

We especially highlight that Seedream 3.0 can directly generate images with higher resolution, like 2048$\times$2048, further enhancing portrait texture. Some examples of Seedream 3.0 can be found in Figure~\ref{fig:human_showcase}. The quality of generated portraits shows promising progress toward professional photography standards, bringing significant new possibilities for the application.

\subsection{Comparison with GPT-4o}
\begin{figure*}[!t]
\centering
\includegraphics[width=\linewidth]{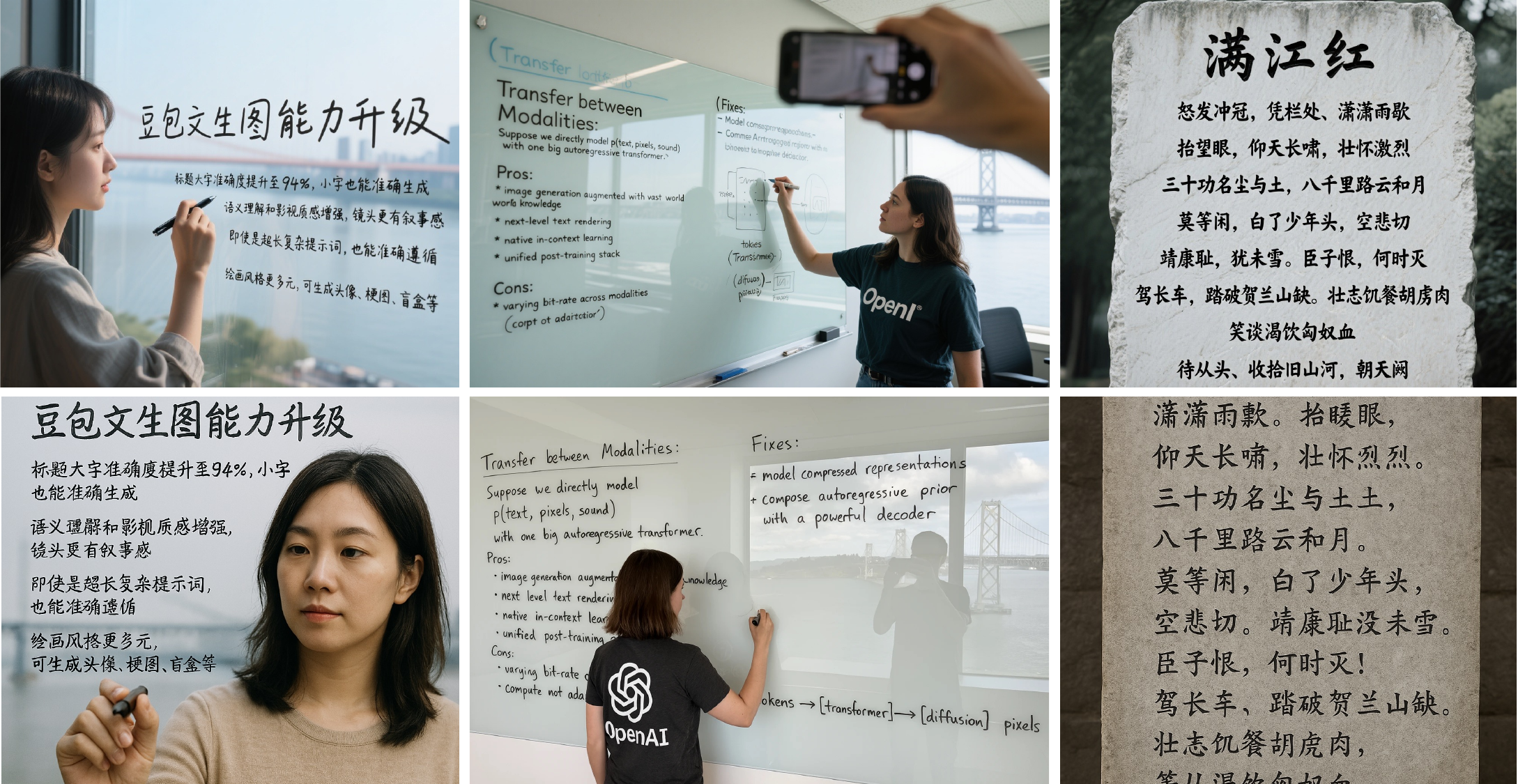}
\caption{Comparisons of Text Rendering. Top for Seedream 3.0 and bottom for GPT-4o. Better to zoom in for better view.}
\label{fig:gpt_comp_text}
\end{figure*}
\begin{figure*}[!t]
\centering
\includegraphics[width=\linewidth]{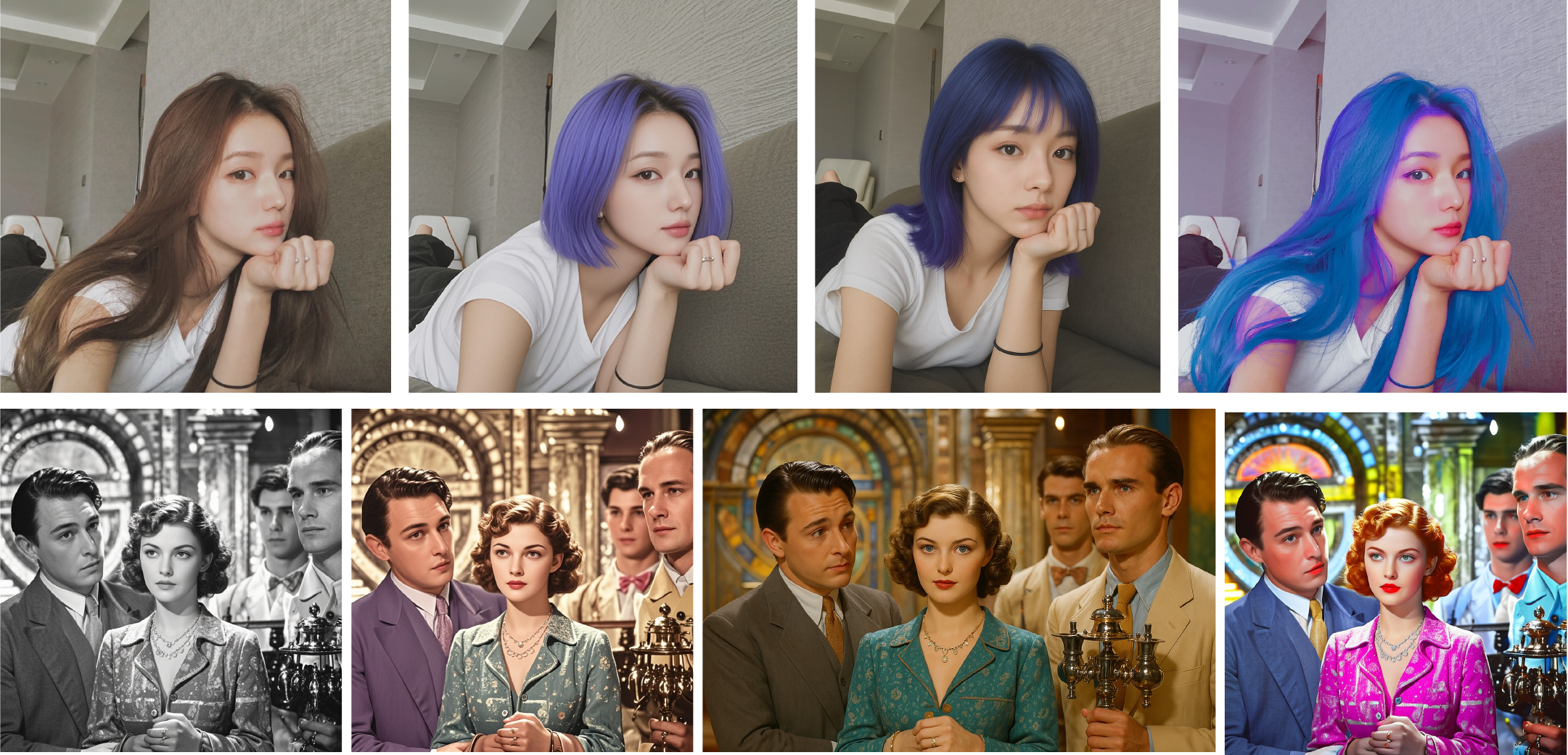}
\caption{Comparisons of Image Edit. From left to right: the original image, SeedEdit 1.6, GPT-4o, and Gemini-2.0. \textbf{Top Prompt}: 换个蓝紫色短发. \textbf{Bottom Prompt}: 变成彩色图片.}
\label{fig:gpt_comp_edit}
\end{figure*}
\begin{figure*}[!t]
\centering
\includegraphics[width=\linewidth]{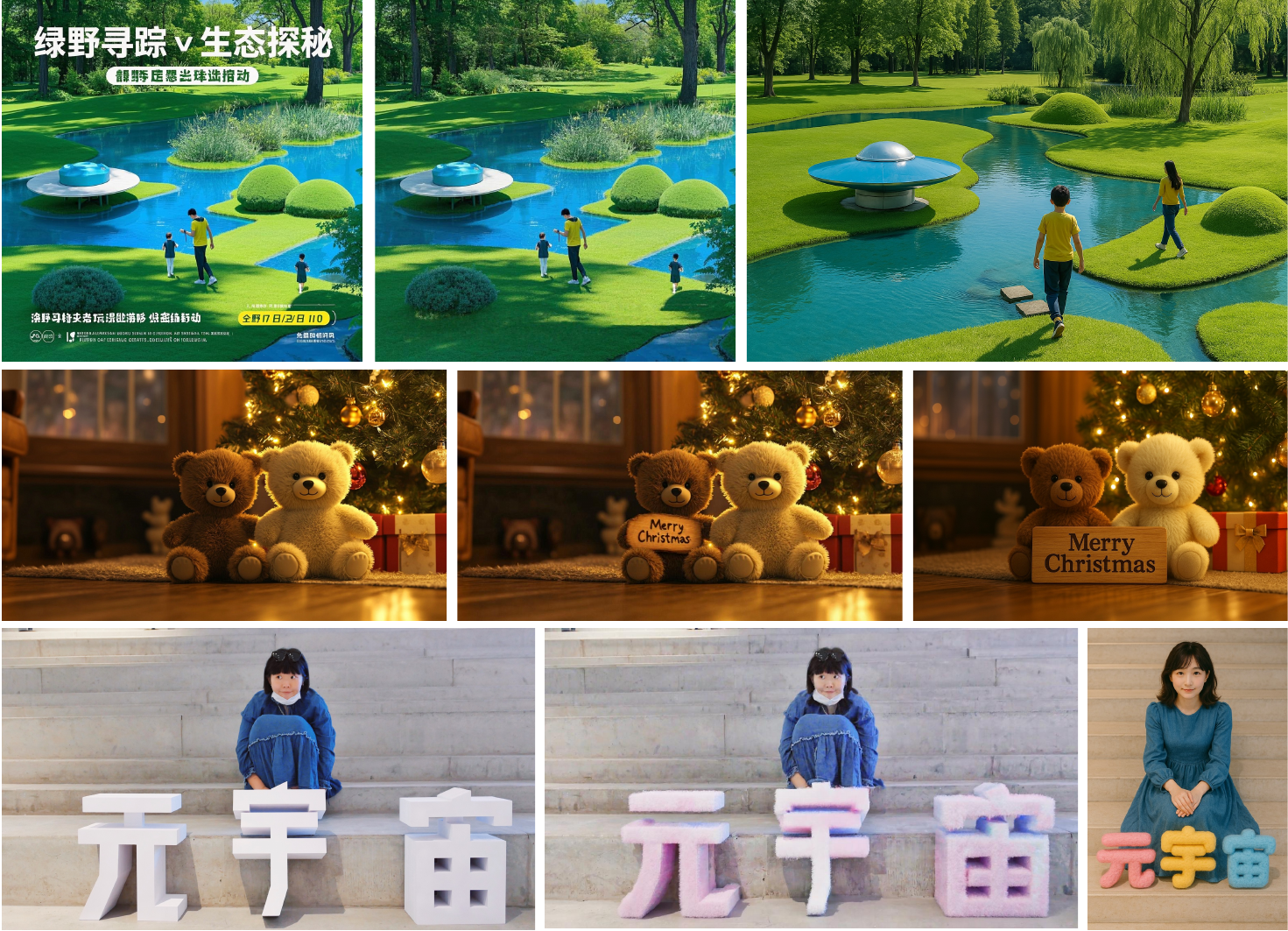}
\caption{Comparisons of Text Edit. From left to right: the original image, SeedEdit, and GPT-4o. \textbf{Top Prompt}: 不要文字. \textbf{Middle Prompt}: 小熊的身前摆了一个小木牌，上面雕刻着"Merry Christmas". \textbf{Bottom Prompt}: 把字改成彩色毛绒材质.}
\label{fig:edit_text}
\vspace{-10pt}
\end{figure*}

Recently, GPT-4o has introduced an impressive image generation function, which features exceptionally powerful multi-modal capabilities. Due to the absence of an API for large-scale image generation, a systematic evaluation has not yet been conducted. Nevertheless, a comparative analysis of selected cases reveals that GPT-4o and Seedream 3.0 each exhibit distinct strengths and weaknesses across different scenarios.

\subsubsection{Dense Text Rendering}
GPT-4o~\cite{gpt-4o} presents impressive text rendering capabilities, as evidenced by multiple examples. We generate comparable cases for comparison, as shown in Figure~\ref{fig:gpt_comp_text}. GPT-4o excels in the accuracy of rendering small English characters and certain LaTeX symbols. However, it exhibits notable limitations in rendering Chinese fonts. In contrast, Seedream 3.0 handles dense Chinese text generation with ease and outperforms GPT-4o in terms of typesetting and aesthetic composition. 

\begin{figure*}[!t]
\centering
\includegraphics[width=\linewidth]{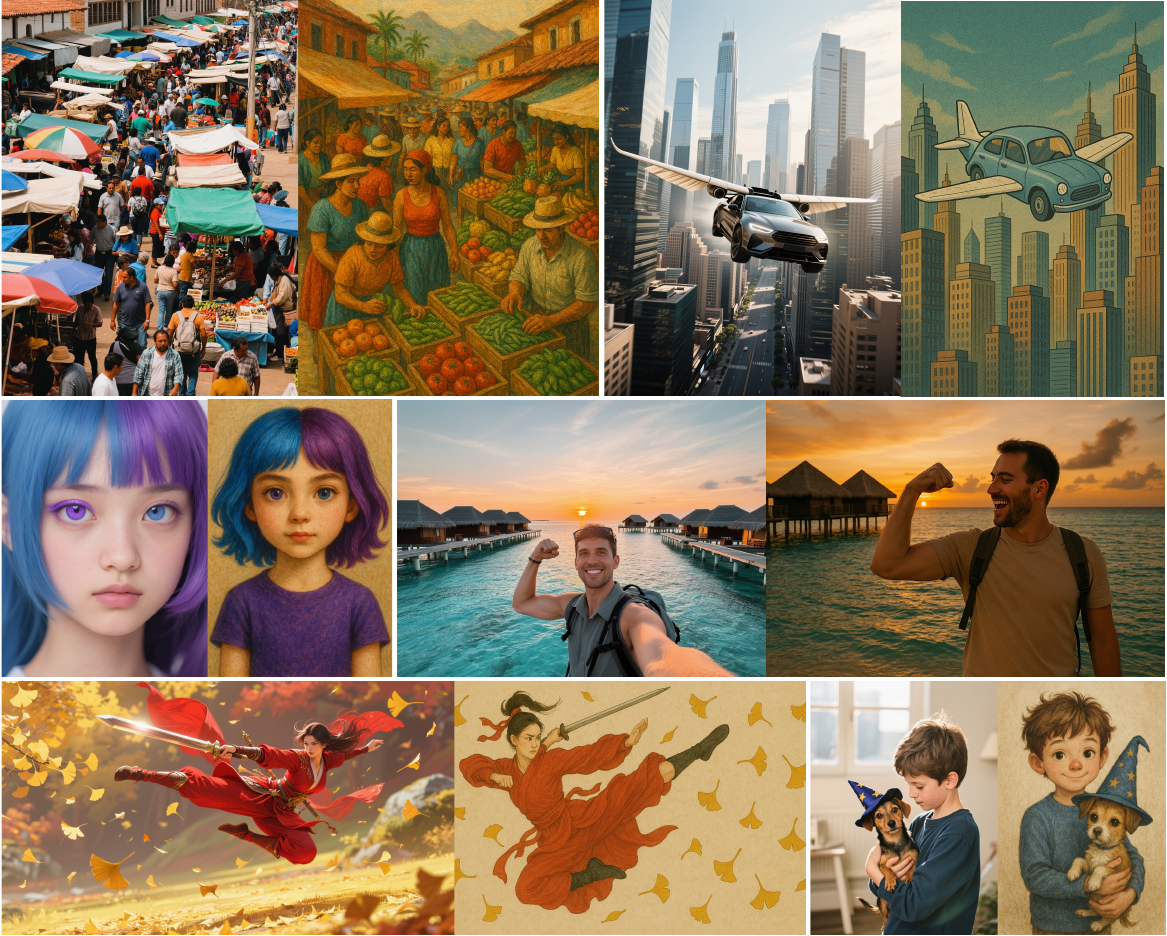}
\caption{Image Quality Comparisons. Left: Seedream 3.0, Right: GPT-4o.}
\label{fig:gpt_comp_quality}
\end{figure*}

\subsubsection{Image Editing}
Image editing tasks bridge the generation with real-world images, attracting significant attention for practical usage. GPT-4o can perform editing operations on given images based on prompt descriptions. SeedEdit, derived from Seedream, also supports such capabilities. Additionally, Gemini-2.0 recently demonstrates strong multi-modal image generation, particularly in interleaved generation and multi-round editing. This study focuses on comparing the single-round image generation capabilities of these models, as shown in Figure~\ref{fig:gpt_comp_edit}. We demonstrate that SeedEdit exhibits better ID preserving and prompt following abilities.

These three models exhibit distinct characteristics. GPT-4o excels at fulfilling a wide range of editing requirements but tends to struggle with preserving the original image, particularly regarding IP and ID consistency. Gemini-2.0 maintains the original image at the pixel level, but often produces issues with color naturalness and image quality. SeedEdit 1.6 provides balanced performance, effectively addressing typical editing needs while maintaining a relatively high availability rate. However, it still faces limitations when handling more complex tasks, such as multi-image reference and multi-round editing. These areas will be improved in future versions.

We primarily compared the performance of SeedEdit and GPT-4o on text-related editing tasks. Text editing is inherently challenging, as it requires not only text rendering but also the recognition and understanding of characters within images. The ability to handle text editing tasks marks a significant advancement in controllable image generation, particularly for real images. Figure~\ref{fig:edit_text} illustrates examples of tasks such as text writing, removing, and modification. SeedEdit inherits the text-related capabilities of Seedream 3.0, delivering satisfying results. It can detect text in images accurately, allowing for precise deletion or modification. Additionally, when adding text, SeedEdit considers the layout and integrates the text seamlessly into the original image. In contrast, while GPT-4o can fulfill text editing requirements, it fails to preserve the original image, limiting its practical use. 

\subsubsection{Generation Quality}

Generation quality, including color, texture, clarity, and aesthetic appeal, is a critical factor in assessing text-to-image models. Seedream models have consistently demonstrated strong performance in these areas, while GPT-4o shows some shortcomings. As shown in Figure~\ref{fig:gpt_comp_quality}, images generated by GPT-4o tend to have a dark yellowish hue and exhibit significant noise, which notably impacts the usability of the generated images in various scenarios.


\section{Conclusion}

In this paper, we have introduced Seedream 3.0, which employs several innovative strategies to address existing challenges, including limited image resolutions,  complex attributes adherence, fine-grained typography generation, and suboptimal visual aesthetics and fidelity. Through system-level upgrades in data construction, model pretraining, post-training, and model acceleration, Seedream 3.0 has achieved comprehensive improvements in multiple aspects compared to our previous version. Seedream 3.0 provides native high-resolution output, comprehensive capability, superior text rendering quality, enhanced visual appeal, and extreme generation speed. With its integration into platforms like Doubao and Jimeng, Seedream 3.0 exhibits strong potential to become a powerful productivity tool across various work and daily life scenarios.

%% file: sections/appendix.tex
\section{Contributions and Acknowledgments}
\label{contributions}

All contributors of Seedream are listed in alphabetical order by their last names.

\subsection{Core Contributors}
        Yu Gao, 
        Lixue Gong, 
	Qiushan Guo,
	Xiaoxia Hou,
        Weilin Huang, 
	Zhichao Lai,
	Fanshi Li,
	Liang Li,
	Xiaochen Lian,
        Chao Liao,
	Liyang Liu,
	Wei Liu,
	Yichun Shi,
	Shiqi Sun,
	Yu Tian,
	Zhi Tian,
	Peng Wang,
	Rui Wang,
	Xuanda Wang,
	Xun Wang,
	Ye Wang,
	Guofeng Wu,
	Jie Wu,
	Xin Xia,
	Xuefeng Xiao, 
        Jianchao Yang, 
	Zhonghua Zhai,
	Xinyu Zhang,
	Qi Zhang,
	Yuwei Zhang,
	Shijia Zhao.

\subsection{Contributors}
	Haoshen Chen, 	
	Kaixi Chen, 
	Xiaojing Dong, 	
	Jing Fang, 	
	Yongde Ge, 	
	Meng Guo, 	
	Shucheng Guo, 	
	Bibo He,   
	Lurui Jin, 	
	Bo Li, 	
	Hao Li,  
	Huixia Li, 	
	Jiashi Li,	
	Ying Li, 
	Yiying Li, 	
	Yameng Li, 	
	Heng Lin, 	
	Feng Ling, 	
	Shu Liu, 	
	Zuxi Liu, 	
	Yanzuo Lu,   
	Wei Lu,   
	Tongtong Ou, 	
	Ke'er Qin, 	
	Yinuo Wang, 
        Yonghui Wu,
	Yao Yao,	
	Fengxuan Zhao,
        Wenliang Zhao,
        Wenjia Zhu.